\documentclass{tlp}
\usepackage{aopmath}
\usepackage{graphicx}
\usepackage{subfigure}
\usepackage{amssymb}
\usepackage[noend]{algpseudocode}
\usepackage{algorithm}

\newtheorem{definition}{Definition} 
\newtheorem{example}{Example} 

\newcommand{\argmax}{\mathit{argmax}} 

\submitted{26 June 2012}
\revised{04 January 2013}
\accepted{07 March 2013}

\begin{document}
\bibliographystyle{acmtrans}

\long\def\comment#1{}

\title[Inference and Learning in PLP using Weighted Formulas]{Inference and Learning in Probabilistic Logic Programs using Weighted Boolean Formulas}

\author[D.~Fierens et al.]
{DAAN FIERENS, GUY VAN DEN BROECK, JORIS RENKENS, \authorbreak DIMITAR SHTERIONOV, BERND GUTMANN, INGO THON, \authorbreak  GERDA JANSSENS, LUC DE RAEDT \\
Department of Computer Science \\  
KU Leuven \\
Celestijnenlaan 200A, 3001 Heverlee, Belgium\\
\email{FirstName.LastName@cs.kuleuven.be}
}

\pagerange{\pageref{firstpage}--\pageref{lastpage}}
\setcounter{page}{1}

\volume{}
\jdate{2}
\pubyear{2013}

\maketitle

\label{firstpage} 

\begin{abstract}
Probabilistic logic programs are logic programs in which some of the
facts are annotated with probabilities.  
This paper investigates how classical inference and learning tasks known
from the graphical model community can be tackled for probabilistic logic programs.
Several such tasks such as computing the marginals given evidence and learning
from (partial) interpretations have not really been addressed for probabilistic logic programs
before. 


The first contribution of this paper is a suite of efficient algorithms for various inference tasks. It is based on a conversion of the program and the queries and evidence to a weighted Boolean formula. This
allows us to reduce the inference tasks to well-studied tasks such as
weighted model counting, which can be solved using state-of-the-art methods known from the graphical model and knowledge compilation literature.   The second contribution is an algorithm for parameter estimation in the learning from interpretations setting. The algorithm employs Expectation Maximization, and is built on top of the developed inference algorithms.

 The proposed approach is experimentally evaluated. The results show that the inference algorithms improve upon the state-of-the-art in probabilistic logic programming and that it is indeed possible to learn the parameters of a probabilistic logic program from interpretations.
\end{abstract}


\section{Introduction}
\label{sec:introduction}

There is a lot of interest in combining probability and logic for dealing with complex relational domains. This interest has resulted in  the fields of \emph{Probabilistic Logic Programming (PLP)} \cite{PILPbook} and \emph{Statistical Relational Learning (SRL)} \cite{SRLbook}. While the two fields essentially study the same problem,
there are differences in emphasis. SRL techniques have focussed on the
extension of probabilistic graphical models like Markov or Bayesian networks with logical and relational representations, as in for instance Markov logic \cite{MCSAT}. Conversely, PLP has extended logic programming languages
(or Prolog) with probabilities. This has resulted in differences in representation and semantics
between the two approaches and, more importantly, also in differences in the inference tasks and learning settings that are supported. In graphical models and SRL, the most common \emph{inference} tasks are that of computing the marginal probability of a set of random variables given some evidence (we call this the MARG task) and finding the most likely joint state of the random variables given the evidence (the MPE task). The PLP community has mostly focussed on computing the success probability of queries without evidence. Furthermore, probabilistic logic programs are usually \emph{learned} from entailment \cite{PRISM_PILPbook,GutmannECML08}, while the standard learning setting in graphical models and SRL corresponds to learning from interpretations. This paper bridges the gap between the two communities, by adapting the traditional graphical model and SRL settings towards the PLP perspective. 
 We contribute general MARG and MPE inference techniques and a learning from interpretations algorithm for PLP. 
In this paper we use ProbLog \cite{ProbLogIJCAI07} as the PLP language, but our approach is relevant to related languages like ICL \cite{ICL_PILPbook}, PRISM \cite{PRISM_PILPbook} and LPAD/CP-logic \cite{VennekensTPLP09} as well.

The first key contribution of this paper is a two-step approach for performing MARG and MPE inference in probabilistic logic programs. In the first step,  the program is converted to an equivalent weighted Boolean (propositional) formula. 
This conversion is based on well-known conversions from the knowledge representation and logic programming literature. 
The MARG task then reduces to 
weighted model counting (WMC) on the resulting weighted formula, and the MPE task to weighted MAX-SAT.   
The second step then involves calling a state-of-the-art solver for WMC or MAX-SAT.
In this way, we establish new links between PLP inference and standard problems such as WMC and MAX-SAT. 
We also identify a novel connection between PLP and Markov Logic \cite{MCSAT}. From a probabilistic perspective, our approach is similar to the work of Darwiche~\shortcite{DarwicheBook} and others \cite{Sang05,Park02}, who perform  Bayesian network inference by conversion to weighted formulas. We do the same for PLP, a much more expressive representation framework than traditional graphical models. 
PLP extends a programming language and allows us to concisely represent large sets of dependencies between random variables.
From a logical perspective, our approach is related to Answer Set Programming (ASP), where models are often computed by translating the ASP program to a Boolean formula and applying a SAT solver \cite{ASSATjrnl04}. Our approach is similar in spirit, but is different in that it employs a probabilistic context.

The second key contribution of this paper is an algorithm for learning the parameters of probabilistic logic programs from data. We use the learning from 
interpretations (LFI) setting, which is the standard setting in graphical models and SRL (although they use different terminology). This setting has also received a lot of attention in inductive logic programming \cite{DeRaedt08}, but has not yet been used for \emph{probabilistic} logic programs. Our algorithm, called LFI-ProbLog, is based on Expectation-Maximization (EM) and is built on top of the inference techniques presented in this paper.



The present paper is based on and integrates our previous papers \cite{FierensUAI11,GutmannECML11} in which inference and learning were studied and implemented separately. Historically, the learning from interpretations approach as detailed by \citeN{GutmannECML11} and \citeN{LFItechreport} was developed first and used  BDDs for inference and learning. The use of BDDs for learning in an 
EM-style is related to the approach of \citeN{Ishihata_ILP2008}, who developed an EM algorithm for propositional BDDs and suggested that their approach can be used to perform learning from entailment for PRISM programs. \citeN{FierensUAI11} later showed that an alternative approach to inference - that is more general, efficient and principled - can be realized using weighted model counting 
and compilation to d-DNNFs rather than  BDDs as in the initial ProbLog implementation \cite{KimmigTPLP11}. The present paper employs the approach by Fierens et al.\ also for learning from interpretations in an EM-style and thus integrates
the two earlier approaches.  The resulting techniques are integrated in a novel implementation, called ProbLog2. While the first ProbLog implementation  \cite{KimmigTPLP11} was tightly integrated in the YAP Prolog engine and employed BDDs, 
ProbLog2 is much closer in spirit to some Answer Set Programming systems than to Prolog and it employs d-DNNFs and weighted model counting.

This paper is organized as follows. We first review the necessary background (Section~\ref{sec:background}) and introduce PLP (Section~\ref{sec:PLP}). Next we state the inference tasks that
 we consider (Section~\ref{sec:tasks}). Then we introduce our two-step
 approach for inference (Section~\ref{sec:conversion} 
and~\ref{sec:inference_section}), and introduce the new learning algorithm (Section~\ref{sec:lfi}). Finally we briefly discuss the implementation of the new system (Section~\ref{sec:implementation}) and evaluate the entire approach by means of experiments on relational data (Section~\ref{sec:experiments}). 



\section{Background}
\label{sec:background}
We now review the basics of first-order logic (FOL) and logic programming (LP). 
Readers familiar with FOL and LP can safely skip this section.

\subsection{First-Order Logic (FOL)}

A \emph{term} is a variable, a constant, or a functor applied to terms. An \emph{atom} is of the form $p(t_1,\ldots ,t_n)$ where $p$ is a predicate of arity $n$ and the $t_i$ are terms. A \emph{formula} is built out of atoms using universal and existential quantifiers and the usual logical connectives $\lnot$, $\lor$, $\land$, $\rightarrow$  and $\leftrightarrow$. A \emph{FOL theory} is a set of formulas that implicitly form a conjunction. An expression is called \emph{ground} if it does not contain variables. A ground (or propositional) theory is said to be in \emph{conjunctive normal form (CNF)} if it is a conjunction of disjunctions of literals. A \emph{literal} is an atom or its negation. Each disjunction of literals is called a \emph{clause}. A disjunction consisting of a single literal is called a \emph{unit clause}. Each ground 
theory can be written in CNF form. 

The \emph{Herbrand base} of a FOL theory is the set of all ground atoms constructed using the predicates, functors and constants in the theory. A Herbrand interpretation, also called a \emph{(possible) world}, 
is an assignment of a truth value to all atoms in the Herbrand base. 
A world or interpretation is called a \emph{model} of the theory if it satisfies all formulas in the theory (in other words, if all formulas evaluate to true in that world).

\subsection{Logic Programming (LP)}
\label{sec:LP}

Syntactically, a normal logic program, or briefly \emph{logic program (LP)} is a set of rules. 
A \emph{rule} (also called a \emph{normal clause}) is a universally quantified expression of the form  \verb$h :- b1, ... , bn$, where $h$ is an atom and $b_1, \ldots, b_n$ are literals. 
The atom $h$ is called the \emph{head} of the rule and $b_1,\ldots , b_n$ the \emph{body}, representing the conjunction $b_1 \land \ldots \land b_n$. A \emph{fact} is a rule that has $\mathit{true}$ as its body and is written more compactly as \verb$h$.

We use the \emph{well-founded semantics} for LPs \cite{wellfounded91}. In the case of a negation-free LP (or \emph{definite program}), the well-founded model is identical to the well-known \emph{Least Herbrand Model (LHM)}. The LHM is 
equal to the least 
of all models obtained when interpreting the LP as a FOL theory of implications. The \emph{least} model is the model that is a subset of all other models (in the sense that it makes the fewest atoms true). Intuitively, the LHM is the set of  all ground atoms that are entailed by the LP. For negation-free LPs, the LHM is guaranteed to exist and be unique. For LPs with negation, we use the well-founded model. We refer to~\citeN{wellfounded91} for details. 
The ProbLog semantics requires all considered logic programs to have a two-valued well-founded model (see Section~\ref{sec:semantics}). For such programs, the well-founded model is identical to the stable model \cite{wellfounded91}.


Intuitively, the reason why one considers only the \emph{least} model of an LP is that LP semantics makes the \emph{closed world assumption (CWA)}. Under the CWA, everything that is not implied to be true is assumed to be false. This has implications on how to interpret rules. Given a ground LP and an atom $a$, the set of all rules with $a$ in the head should be read as the \emph{definition} of $a$: the atom $a$ is defined to be true if and only if at least one of the rule bodies is true (the `only if' is due to the CWA). This means that there is a crucial difference in semantics between LP and FOL since FOL does not make the CWA. For example, the FOL theory $\{a \leftarrow b\}$ has 3 models $\{\lnot a, \lnot b\}$, $\{a, \lnot b\}$ and $\{a, b\}$. The LP $\{$\verb;a :- b;$\}$ has only one model, namely the least Herbrand model $\{ \lnot a, \lnot b \}$ (intuitively, $a$ and $b$ are false because there is no rule that makes $b$ true, and hence there is no applicable rule that makes $a$ true either). 

Because of the syntactic restrictions of LP, it is tempting to believe that FOL is more `expressive' than LP. This is wrong because of the difference in semantics: certain concepts that can be expressed in 
LP cannot be expressed in 
FOL (see Section~\ref{sec:relatedlang} for details). 
This motivates our interest in LP and PLP.

\section{Probabilistic Logic Programming and ProbLog}
\label{sec:PLP}
Most probabilistic logic programming languages, including PRISM \cite{PRISM_PILPbook}, ICL \cite{ICL_PILPbook}, ProbLog \cite{ProbLogIJCAI07} and LPAD \cite{VennekensTPLP09},
are based on Sato's \emph{distribution semantics} \cite{SatoICLP95}. 
In this paper we use ProbLog, but our
approach can be used for the other languages as well.

\subsection{Syntax of ProbLog}

A ProbLog program consists of two parts: a set of ground probabilistic facts, 
and a logic program, i.e.\ a set of rules and (`non-probabilistic') facts. A ground \emph{probabilistic
  fact}, written \verb$p::f$,  is a ground fact \verb$f$ annotated with a
probability \verb$p$. We allow syntactic sugar for compactly specifying an entire set of probabilistic facts with a single statement. Concretely, we allow what we call \emph{intensional} probabilistic facts, which are statements of the form \verb$p::f(X1,X2,...,Xn) :- body$, with \verb$body$ a conjunction of calls to non-probabilistic facts.\footnote{The notion of intensional probabilistic facts does not appear in earlier ProbLog papers but is often useful in practice.} The idea is that such a statement defines the domains of the variables {\tt X1, X2, ... } and {\tt Xn}. When defining the semantics, as well as when performing inference or learning, an intensional probabilistic fact should be replaced by its corresponding set of ground probabilistic facts, as illustrated below. An atom that unifies with a ground probabilistic fact is called a \emph{probabilistic atom}, while an atom that unifies with the head of some rule in the logic program is called a \emph{derived atom}. The set of probabilistic atoms must be disjoint from the set of derived atoms. Also, the rules in the program should be range-restricted: all variables in the head of a rule should also appear in a positive literal in the body of the rule.

Our running example is the program that models the well-known `Alarm' Bayesian network.
\begin{example}[Running Example]
\label{example:alarm}
\begin{verbatim}
0.1::burglary.                         person(mary).
0.2::earthquake.                       person(john).     
0.7::hears_alarm(X) :- person(X).    
alarm :- burglary.
alarm :- earthquake.
calls(X) :- alarm, hears_alarm(X).
\end{verbatim}
This Problog program consists of probabilistic facts and a
logic program. Predicates of probabilistic atoms are \verb$burglary/0$,
\verb$earthquake/0$ and \verb$hears_alarm/1$, predicates of derived atoms are
\verb$person/1$, \verb$alarm/0$ and \verb$calls/1$. 
Intuitively, the probabilistic facts \verb$0.1::burglary$ and
\verb$0.2::earthquake$  state that there is a burglary with
probability 0.1 and  an earthquake with probability 0.2.
The statement \verb$0.7::hears_alarm(X) :- person(X)$ is an intensional probabilistic fact and is syntactic sugar for the following set of ground probabilistic facts.

\begin{verbatim}
0.7::hears_alarm(mary).
0.7::hears_alarm(john).
\end{verbatim}

The rules in the program define 
when the alarm goes off and when a person calls, as a function of the probabilistic facts.
\end{example}

\subsection{Semantics of ProbLog}
\label{sec:semantics}

A ProbLog program specifies a probability
distribution over possible worlds. To define this distribution, it is
easiest to consider the grounding of the program  with respect to the
Herbrand base.\footnote{Beforehand, a  preprocessing step already replaced the intensional probabilistic  facts with their corresponding ground set, as illustrated before.} In this paper, we assume that the resulting Herbrand base is finite. For the distribution semantics in the infinite case, see \citeN{SatoICLP95}. 

Each ground probabilistic fact \verb$p::f$ gives an
\emph{atomic choice}, i.e.\ we can choose to include $f$ as a fact
(with probability $p$) or discard it (with probability $1-p$). A
\emph{total choice} is obtained by making an atomic choice for each
ground probabilistic fact. Formally, a total choice is any subset
of the set of all ground probabilistic atoms. Hence, if there are $n$
ground probabilistic atoms then there are $2^n$ total
choices. Moreover, we have a probability distribution over these total
choices: the probability of a total choice is defined to be the
product of the probabilities of the atomic choices that it is composed
of (we can take the product since atomic choices are seen as independent events). 

\begin{example}[Total Choices of the Alarm Example]
\label{example:alarm:choices}
Consider the Alarm program of Example~\ref{example:alarm}.
The $2^4=16$ total
choices corresponding to the 4 ground probabilistic atoms are given in Table~\ref{table:alarm:pw}.
The first row corresponds to the total choice in which all the
probabilistic atoms are true. 
The probability of this total choice is 0.1 $\times$ 0.2 $\times$ 0.7 $\times$
0.7 = 0.0098. The second row corresponds to the same total choice except that $\mathit{hears\_alarm}(\mathit{mary})$ is now false. 
The probability is 0.1 $\times$ 0.2 $\times$ 0.7 $\times$ (1-0.7) = 0.0042. The sum of probabilities of all 16 total choices is equal to one.
\end{example} 
\begin{table}[htb]
  \caption{Total choices and their probabilities}
  \small
  \begin{tabular}{r|l|r}
& Total choice $C$ & P(C) \\
\hline
 1 & $\{$  $\mathit{burglary},$  $\mathit{earthquake},$  $\mathit{hears\_alarm}(\mathit{john}),$  $\mathit{hears\_alarm}(\mathit{mary})$  $\}$ & 0.0098\\
2 & $\{$  $\mathit{burglary},$  $\mathit{earthquake},$  $\mathit{hears\_alarm}(\mathit{john})$    $\}$ & 0.0042 \\
3 & $\{$  $\mathit{burglary},$  $\mathit{earthquake},$    $\mathit{hears\_alarm}(\mathit{mary})$  $\}$ & 0.0042 \\
4 & $\{$  $\mathit{burglary},$  $\mathit{earthquake}$      $\}$ & 0.0018 \\

5 & $\{$  $\mathit{burglary},$    $\mathit{hears\_alarm}(\mathit{john}),$  $\mathit{hears\_alarm}(\mathit{mary})$  $\}$ & 0.0392 \\
6 & $\{$  $\mathit{burglary},$    $\mathit{hears\_alarm}(\mathit{john})$    $\}$ & 0.0168 \\
7 & $\{$  $\mathit{burglary},$      $\mathit{hears\_alarm}(\mathit{mary})$  $\}$ & 0.0168 \\
8 & $\{$  $\mathit{burglary}$        $\}$ & 0.0072 \\

9 & $\{$    $\mathit{earthquake},$  $\mathit{hears\_alarm}(\mathit{john}),$  $\mathit{hears\_alarm}(\mathit{mary})$  $\}$ & 0.0882 \\
10 & $\{$    $\mathit{earthquake},$  $\mathit{hears\_alarm}(\mathit{john})$    $\}$ & 0.0378 \\
11 & $\{$    $\mathit{earthquake},$    $\mathit{hears\_alarm}(\mathit{mary})$  $\}$ & 0.0378 \\
12 & $\{$    $\mathit{earthquake}$      $\}$ & 0.0162 \\

13 & $\{$      $\mathit{hears\_alarm}(\mathit{john}),$  $\mathit{hears\_alarm}(\mathit{mary})$  $\}$ & 0.3528 \\
14 & $\{$      $\mathit{hears\_alarm}(\mathit{john})$    $\}$ & 0.1512 \\
15 & $\{$        $\mathit{hears\_alarm}(\mathit{mary})$  $\}$ & 0.1512 \\
16 & $\{$          $\}$ & 0.0648 \\
\hline
\end{tabular}
\normalsize
  \label{table:alarm:pw}
\end{table}

Given a particular total choice $C$, we obtain a logic program $C \cup R$, where $R$ denotes the rules in the ProbLog program. We denote the well-founded model of this logic program as $WFM(C \cup R)$.\footnote{Recall from Section~\ref{sec:LP} that for negation-free programs, the WFM is the least Herbrand model.} We call a given world $\omega$ a \emph{model} of the ProbLog program if there indeed exists a total choice $C$ such that $WFM(C \cup R) = \omega$. We use $MOD(L)$ to denote the set of all models of a ProbLog program $L$. The ProbLog semantics is only well-defined for programs that are \emph{sound} \cite{RiguzziTPLP12}, i.e., programs for which each possible total choice $C$ leads to a well-founded model that is two-valued or `total' \cite{RiguzziTPLP12,wellfounded91}.\footnote{A sufficient condition for this is that the rules in the ProbLog program are \emph{locally stratified} \cite{wellfounded91}. In particular, this trivially holds for all negation-free programs.}  Programs for which this is not the case are not considered valid ProbLog programs.


Everything is now in place to define the distribution over possible
worlds: the probability of a world that is a model of the ProbLog
program is equal to the probability of its total choice; the
probability of a world that is not a model is 0. 

\begin{example}[Models and their probabilities]
\label{example:alarm:models}
(Continuing Example~\ref{example:alarm:choices}) The total choice $\{\mathit{burglary},\mathit{earthquake},\mathit{hears\_alarm}(\mathit{john})\}$, which has probability 0.1 $\times$ 0.2 $\times$ 0.7 $\times$ (1-0.7) = 0.0042, yields the following logic program.
\begin{verbatim}
burglary.                    person(mary).             
earthquake.                  person(john).             
hears_alarm(john).        
alarm :- earthquake.
alarm :- burglary.
calls(X) :- alarm, hears_alarm(X).
\end{verbatim}
The WFM of this program is the world
$\{\mathit{person}(\mathit{mary}), \mathit{person}(\mathit{john}), \mathit{burglary},$ $\mathit{earthquake},\mathit{hears\_alarm}(\mathit{john}), \lnot \mathit{hears\_alarm}(\mathit{mary}), \mathit{alarm}, \mathit{calls}(\mathit{john}), \lnot \mathit{calls}(\mathit{mary}) \}$. 
Hence this world is a model and its probability is 0.0042.
In total there are 16 models, corresponding to each of the 16 total choices shown in Table~\ref{table:alarm:pw}. Note that, out of all possible interpretations of the vocabulary, there are many that are not models of the ProbLog program. An example is  any world of the form $\{\mathit{burglary}, \lnot \mathit{alarm}, \ldots\}$: it is impossible that $\mathit{alarm}$ is false while $\mathit{burglary}$ is true. The probability of such worlds is zero.
\end{example}


\subsection{Related Languages}
\label{sec:relatedlang}

ProbLog is strongly related to several other languages, in particular to Probabilistic Logic Programming (PLP) languages like PRISM \cite{PRISM_PILPbook}, ICL \cite{ICL_PILPbook} and LPAD \cite{VennekensTPLP09}, and other languages like Markov Logic \cite{MCSAT}. Table~\ref{table:languages} shows the main features of each language and the major corresponding system.  

\begin{table}[ht]
  \caption{Overview of features of several probabilistic logical languages and the corresponding systems (implementations). The first three features are properties of the language, the last two are properties of the system. We refer to the first ProbLog system as ProbLog1 and to the system described here as ProbLog2.}
  \small
  \begin{tabular}{lcccccc}
\hline
Language & ProbLog & ProbLog & PRISM & ICL & LPAD & MLN \\
System & ProbLog1 & ProbLog2 & PRISM & AILog2 & PITA & Alchemy \\
\hline
Cyclic rules             & $\checkmark$ & $\checkmark$ & $-$          & $-$          & $\checkmark$ & $\checkmark$ \\
{\parbox{0.20\textwidth}{\vspace{0.12cm}Overlapping\\ rule bodies\vspace{0.12cm}}}  & $\checkmark$ & $\checkmark$ & $-$          & $\checkmark$ & $\checkmark$ & n/a \\
{\parbox{0.20\textwidth}{\vspace{0.12cm}Inductive \\ definitions\vspace{0.12cm}}}    & $\checkmark$ & $\checkmark$ & $\checkmark$ & $\checkmark$ & $\checkmark$ & $-$ \\
\hline
{\parbox{0.20\textwidth}{Evidence on\\ arbitrary atoms\vspace{0.12cm}}} & $-$          & $\checkmark$ & $-$          & $\checkmark$          & $-$          & $\checkmark$ \\
Multiple queries         & $-$          & $\checkmark$ & $-$          & $-$          & $-$          & $\checkmark$ \\
\hline
  \end{tabular}
\normalsize
  \label{table:languages}
\end{table}

Compared to most other PLP languages, ProbLog is more expressive with respect to the rules that are allowed in a program. This holds in particular for PRISM and ICL. Both PRISM and ICL require the rules to be acyclic (or
contingently acyclic) \cite{PRISM_PILPbook,ICL_PILPbook}. In ProbLog we can have cyclic programs with rules such as \verb$smokes(X) :- smokes(Y), influences(Y,X)$. This type of cyclic rules are often needed for tasks such as
collective classification or social network analysis (see Section \ref{sec:experiments}). In addition to acyclicity, PRISM also requires rules with unifiable heads to have mutually exclusive bodies (such
that at most one of these bodies can be true simultaneously; this is the mutual
exclusiveness assumption). ProbLog does not have this restriction, so rules with unifiable heads can have `overlapping' bodies. For instance, the bodies of the two alarm rules in our running example are overlapping: either burglary or earthquake is sufficient for making the alarm go off, but both can also happen at the same time.

LPADs, as used in the PITA system \cite{RiguzziTPLP12}, do not have these syntactic restrictions, and are hence on par with ProbLog in this respect. However, the PITA system does not support the same tasks as the new ProbLog2 system does. For instance, when computing marginal probabilities, ProbLog2 can deal with multiple queries simultaneously and can incorporate evidence, while PITA uses the more traditional PLP setting which considers one query at a time, without evidence (the \emph{succes probability} setting, see Section~\ref{sec:tasks}). The same also holds for the first ProbLog system \cite{KimmigTPLP11}. Note that while evidence can in some special cases be incorporated through modelling,\footnote{For instance, when encoding a Bayesian network in PLP, evidence on nodes at the top of the network (nodes without parents) can be incorporated by including deterministic facts in the program.} we here focus on the general case, i.e., the ability of the system to handle evidence on any arbitrary subset of all atoms in the Herbrand base.

ProbLog2 is the first PLP system that posesses all the features considered in Table~\ref{table:languages}, i.e., that supports multiple queries and evidence while having none of the language restrictions. The experiments in this paper (Section \ref{sec:experiments}) require all these features and can hence only be carried out in ProbLog2, but not in the other PLP systems.

Markov Logic \cite{MCSAT} is strictly speaking not a PLP language as it is based on First-Order Logic instead of Logic Programming. Nevertheless, Markov Logic of course serves the same purpose as the above PLP languages. In terms of expressivity, Markov Logic has the drawback that it cannot express (non-ground) \emph{inductive definitions}. An example of an inductive definition is the definition of the notion of a path in a graph in terms of the edges. This can be written in plain Prolog and hence also in ProbLog. 
\begin{verbatim}
path(X,Y) :- edge(X,Y).
path(X,Y) :- edge(X,Z), path(Z,Y).
\end{verbatim}
In the knowledge representation community, it is well-known that inductive definitions can naturally be represented in Logic Programming (LP), due to LP's least or well-founded model semantics \cite{Denecker01}. In contrast, in First-Order Logic (FOL) one cannot express non-ground inductive definitions, such as the \emph{path} definition above \cite{transitive92}. The reason is, roughly speaking, that \emph{path} is the transitive closure of \emph{edge}, and FOL can express that a given relation is transitive, but cannot in general specify this closure. This result carries over to the probabilistic case: we can express inductive definitions in PLP languages like ProbLog but not in FOL-based languages like Markov Logic.\footnote{This discussion applies to \emph{non-ground} ProbLog programs and Markov Logic Networks (MLNs). In Section~\ref{sec:wbool} we show that every \emph{ground} ProbLog program can be converted to an equivalent ground MLN. The above implies that no such conversion exists on the non-ground (first-order) level.} While the non-probabilistic case has been well-studied in the knowledge representation literature \cite{Denecker01,transitive92}, the probabilistic case has only very recently received attention \cite{Fierens12NIPS}.

\section{Inference Tasks}
\label{sec:tasks}

In the literature on probabilistic graphical models and statistical
relational learning, the two most common inference tasks are 
computing the marginal probability of a set of
random variables given some observations or evidence (we call this the MARG task), and finding the
most likely joint state of the random variables given the evidence (known as the MPE task, for Most Probable Explanation).
In PLP, the focus has been on the special case of MARG where there is only a
single query atom $Q$ 
and no evidence. 
This task is often called computing
the \emph{success probability} of $Q$ \cite{ProbLogIJCAI07}. 
The only works related to the general MARG or MPE task in the PLP literature make a number of restrictive assumptions about the given program such as acyclicity \cite{GutmannECML11} and the mutual exclusiveness assumption of PRISM \cite{PRISM_PILPbook}. 
There also exist approaches that transform ground probabilistic programs to Bayesian networks
and then use  standard Bayesian network inference procedures \cite{Meert09}. However, these are also restricted to acyclic 
and already grounded programs. 

Our approach for the MARG and MPE inference tasks does not suffer from such restrictions and is applicable to all ProbLog programs. We now formally define these tasks, in addition to a third, strongly related task. Let $\mathbf{At}$ be the Herbrand base, i.e, the set of all ground (probabilistic and derived)
atoms in a given ProbLog program. We assume that we are given a set
$\mathbf{E} \subset \mathbf{At}$ of observed atoms and a vector $\mathbf{e}$ with their observed truth values. We refer to this as the \emph{evidence} and write $\mathbf{E}=\mathbf{e}$. Note that the evidence is essentially a partial interpretation of the atoms in the ProbLog program. 
\begin{itemize}
\item In the \textbf{MARG} task, we are given a  set $\mathbf{Q} \subset \mathbf{At}$  
of atoms of interest, called \emph{query atoms}. The task is to compute the marginal probability distribution of every such atom given the evidence, i.e.\ compute $P(Q \mid \mathbf{E}=\mathbf{e})$ for each $Q \in \mathbf{Q}$.\footnote{The common PLP task of computing the \emph{success probability} of an atom $Q$ is a special case of MARG with $\mathbf{Q}$ being the singleton $\{Q\}$ and $\mathbf{E}=\emptyset$.}
\item The \textbf{EVID} or `probability of evidence' task is to compute $P(\mathbf{E}=\mathbf{e})$. It corresponds to the likelihood of data in a learning setting and can be used as a building block for solving the MARG task (see Section~\ref{sec:MARGsolution}).
\item The \textbf{MPE} task is to find the most likely interpretation (joint state) of all non-evidence atoms given the evidence, i.e.\ finding $\argmax_{\mathbf{u}} P( \mathbf{U}=\mathbf{u} \mid \mathbf{E}=\mathbf{e})$, with $\mathbf{U}$ being the unobserved atoms, i.e., $\mathbf{U} = \mathbf{At} \setminus \mathbf{E}$.
\end{itemize}
As the following example illustrates, the different tasks are strongly related.

\begin{example}[Inference tasks]
\label{example:alarm:inference}
Consider the ProbLog program of Example~\ref{example:alarm} and assume that we know that John calls, so $\mathbf{E}=\{\mathit{calls}(\mathit{john})\}$ and $\mathbf{e}=\{\mathit{true}\}$. It can be verified that $\mathit{calls}(\mathit{john})$ is true in 6 of the 16 models of the program, namely the models of total choices 1, 2, 5, 6, 9 and 10 of Table~\ref{table:alarm:pw}. The sum of their
probabilities is 0.196, so this is the probability of evidence (EVID). The MPE task boils down to finding the world with the highest probability out of the 6 worlds that have $\mathit{calls}(\mathit{john}) = \mathit{true}$. It can be verified that this is the world corresponding to total choice 9, i.e., the choice $\{\mathit{earthquake}, \mathit{hears\_alarm}(\mathit{john}), \mathit{hears\_alarm}(\mathit{mary})\}$. An example of the MARG task is to compute the probability that there is a burglary, i.e., $P(\mathit{burglary} = \mathit{true} \mid  \mathit{calls}(\mathit{john}) = \mathit{true} ) = \frac{P(\mathit{burglary} = \mathit{true} \land  \mathit{calls}(\mathit{john}) = \mathit{true})}{P( \mathit{calls}(\mathit{john}) = \mathit{true})}$. There are 4 models in which both $\mathit{calls}(\mathit{john})$ and $\mathit{burglary}$ are true (models 1, 2, 5 and 6), and their sum of probabilities is 0.07. Hence, $P(\mathit{burglary} = \mathit{true}  \mid \mathit{calls}(\mathit{john}) = \mathit{true} )$= 0.07/ 0.196 = 0.357.
\end{example}

Our approach to inference consists of two steps: 1) convert the program to a weighted Boolean formula and 2) perform inference on the resulting weighted formula.
We discuss these two steps in the next sections.


\section{Conversion to a Weighted Formula}
\label{sec:conversion}

Our conversion takes as input a ProbLog program $L$, evidence $\mathbf{E=e}$ and a set of \emph{query atoms} $\mathbf{Q}$, and returns a weighted Boolean (propositional) formula that contains all necessary information. The conversion is similar for each of the considered tasks (MARG, MPE or EVID). The only difference is the choice of the query set $\mathbf{Q}$. For MARG, $\mathbf{Q}$ is the set of atoms for which we want to compute marginal probabilities. For EVID and MPE, we can take $\mathbf{Q} = \emptyset$ (see Section~\ref{sec:inference}). 

The outline of the conversion algorithm is as follows.
\begin{enumerate}
\item \emph{Ground $L$ yielding a program $L_g$ while taking into account  $\mathbf{Q}$ and $\mathbf{E=e}$ (cf.\ Theorem \ref{theorem:rgp}, Section~\ref{sec:RGP}).}\\
It is unnecessary to consider the full grounding of the program, we only need the part that is relevant to the query  given the evidence, that is, the part that captures the distribution $P(\mathbf{Q} \mid \mathbf{E}=\mathbf{e})$. We refer to the resulting program $L_g$ as the \emph{relevant ground program} with respect to $\mathbf{Q}$ and $\mathbf{E}=\mathbf{e}$. 
\item \emph{Convert the ground rules in $L_g$ to an equivalent Boolean formula $\varphi_r$ (cf.\ Lemma \ref{lemma:lp2bool}, Section~\ref{sec:lp2bool}).}\\
This step converts the logic programming rules to an equivalent formula.

\item 
\emph{Assert the evidence and define a weight function (cf.\ Theorem~\ref{theorem:wbool}, Section~\ref{sec:wbool}).}\\ 
To obtain the weighted formula, we first assert the evidence by defining the formula $\varphi$ as the conjunction of the formula $\varphi_r$ for the rules (step2) and for the evidence $\varphi_e$. 
Then we define a \emph{weight function} for all atoms in $\varphi$.
\end{enumerate}
The correctness of the algorithm is shown below; this relies on the indicated theorems and lemma's. 
Before describing the algorithm in detail, we illustrate it on our Alarm example.

\begin{example}[The three steps in the conversion]
\label{example:alarm:steps}
As in Example~\ref{example:alarm:inference}, we take $\mathit{calls}(\mathit{john}) = \mathit{true}$ as evidence. Suppose that we want to compute the marginal probability of $\mathit{burglary}$, so the query set $\mathbf{Q}$ is $\{ \mathit{burglary} \}$. The relevant ground program is as follows. 

\begin{verbatim}
% ground probabilistic facts
0.1::burglary.       0.2::earthquake.        0.7::hears_alarm(john).        
% ground rules
alarm :- burglary.
alarm :- earthquake.
calls(john) :- alarm, hears_alarm(john).
\end{verbatim}
Note that $\mathit{mary}$ does not appear in the grounding because, if we have no evidence about her hearing the alarm or calling, she does not affect the probability $P(\mathit{burglary} \mid \mathit{calls}(\mathit{john}) = \mathit{true})$.

Step  2 converts the three ground rules of the relevant ground program to an
equivalent propositional formula $\varphi_r$ (see Section~\ref{sec:lp2bool}). This formula is the conjunction of $\mathit{alarm} \leftrightarrow \mathit{burglary} \lor \mathit{earthquake}$ and $\mathit{calls}(\mathit{john}) \leftrightarrow \mathit{alarm} \land \mathit{hears\_alarm}(\mathit{john})$.\footnote{For subsequent steps, it is often convenient to write this formula in conjunctive normal form (CNF). For example, some knowledge compilation systems require CNF input.} 
Step 3 adds the evidence. Since we have only one
evidence atom in our example (namely, $\mathit{calls}(\mathit{john})$ is true), all we need to
do is to add the positive unit clause $\mathit{calls}(\mathit{john})$ to the formula $\varphi_r$. The
resulting formula $\varphi$ is $\varphi_r \land \mathit{calls}(\mathit{john}) $. Step 3 also defines the weight function, which assigns a weight 
to each literal in $\varphi$, see Section \ref{sec:wbool}. 
This  results in the \emph{weighted formula}, that is, the combination of the weight function and the Boolean formula $\varphi$. 
\end{example}

We now explain the three steps of the conversion in detail.

\subsection{The Relevant Ground Program}
\label{sec:RGP}

In order to convert the ProbLog program to a Boolean formula we first 
ground it. 
We try to find the part of the grounding that is relevant to the queries $\mathbf{Q}$ and the evidence $\mathbf{E}=\mathbf{e}$. In SRL, this is also called knowledge-based model construction \cite{BLPtechreport}. To do this, we make use of the concept of a dependency set with respect to a ProbLog program. We first explain our algorithm and then show its correctness. 

The \emph{dependency set} of a ground atom $a$ is the set of all ground atoms that occur in some proof of $a$. The dependency set of multiple atoms is the union of their dependency sets. We call a ground atom \emph{relevant} with respect to $\mathbf{Q}$ and $\mathbf{E}$ if it occurs in the dependency set of $\mathbf{Q} \cup \mathbf{E}$. 
We call a ground rule relevant if it contains only relevant atoms. It is safe to restrict the grounding to the relevant rules only. To find the relevant atoms and rules, we apply SLD resolution to prove all atoms in $\mathbf{Q} \cup \mathbf{E}$ (this can be seen as backchaining over the rules starting from $\mathbf{Q} \cup \mathbf{E}$). We employ tabling to avoid proving the same atom twice (and to avoid going into an infinite loop if the rules are cyclic). The relevant rules are all ground rules encountered during the resolution process. As our ProbLog programs are range-restricted, all the variables in the rules used during the SLD resolution will eventually become ground, and hence also the rules themselves.

The above grounding algorithm is not optimal as it does not make use of all available information. For instance, it does not make use of exactly what the evidence is (the values $\mathbf{e}$), but only of which atoms are in the evidence (the set $\mathbf{E}$). One simple, yet sometimes very effective, optimization is to prune \emph{inactive} rules. We call a ground rule inactive if the body of the rule contains  a literal $l$ that is false in the evidence ($l$ can be an atom that is false in $\mathbf{e}$, or the negation of an atom that is true in $\mathbf{e}$). Inactive rules do not contribute to the semantics of a program. Hence they can be omitted. In practice, we do this simultaneously with the above process: we omit inactive rules encountered during the SLD resolution.\footnote{This deals with literals that are \emph{false} in the evidence. Conversely, when a body of a ground rule contains a literal that is \emph{true} in the evidence, it has to be kept and the rule cannot be simplified.  The reason is that the atom's presence might give rise to a positive loop, which has to be detected during the conversion of the ground program to a Boolean formula in the next step.}

The result of this grounding algorithm is what we call the
\emph{relevant ground program} $L_g$ for $L$ with respect to
$\mathbf{Q}$ and $\mathbf{E}=\mathbf{e}$. It contains all the
information necessary for solving the corresponding EVID, MARG or MPE task. 
The advantage of this `focussed' approach (i.e., taking into account $\mathbf{Q}$ and $\mathbf{E}=\mathbf{e}$ during grounding) is
that the program and hence the weighted formula becomes more compact, which makes subsequent
inference more efficient. The disadvantage is that we need to redo the
conversion to a weighted formula when the evidence and queries change. This
is no problem since the conversion is fast compared to the actual
inference (see Section~\ref{sec:experiments}).

The following theorem shows the correctness of our approach.
\begin{theorem} \label{theorem:rgp}
Let $L$ be a ProbLog program and let $L_g$ be the relevant ground program for $L$ with respect to $\mathbf{Q}$ and $\mathbf{E}=\mathbf{e}$. $L$ and $L_g$ specify the same distribution $P(\mathbf{Q} \mid \mathbf{E}=\mathbf{e})$.
\end{theorem}
The proofs of all theorems in this paper are given in the appendix.

We already showed the relevant ground program for the Alarm example in Example~\ref{example:alarm:steps} (in that case, there were irrelevant rules about $\mathit{mary}$, but no inactive rules because there was no negative evidence). To illustrate our approach for cyclic programs, we use the well-known Smokers example \cite{MLN_PILPbook}. 

\begin{example}[ProbLog program for Smokers]
\label{example:smokers:ProbLog}
The ProbLog program for the Smokers example models two causes for people to smoke: either they spontaneously start because of stress or they are influenced by one of their friends.
\begin{verbatim}
0.2::stress(P) :- person(P).                           
0.3::influences(P1,P2) :- friend(P1,P2).            
person(p1).       person(p2).       person(p3).
friend(p1,p2).    friend(p1,p3)
friend(p2,p1).    friend(p3,p1).    

smokes(X) :- stress(X).
smokes(X) :- smokes(Y), influences(Y,X).
\end{verbatim}
With the evidence $\{ \mathit{smokes}(\mathit{p2}) = \mathit{true}, \mathit{smokes}(\mathit{p3}) = \mathit{false}\}$ and the query
set $\{\mathit{smokes}(\mathit{p1})\}$, we obtain the following ground program:
\begin{verbatim}
0.2::stress(p1).         0.2::stress(p2).          0.2::stress(p3).                      
0.3::influences(p2,p1).  0.3::influences(p1,p2).   0.3::influences(p1,p3).        
% irrelevant probabilistic fact !!  0.3::influences(p3,p1).     

smokes(p1) :- stress(p1).
smokes(p1) :- smokes(p2), influences(p2,p1).
% inactive rule !!  smokes(p1) :- smokes(p3), influences(p3,p1).
smokes(p2) :- stress(p2).
smokes(p2) :- smokes(p1), influences(p1,p2).
smokes(p3) :- stress(p3).
smokes(p3) :- smokes(p1), influences(p1,p3).
\end{verbatim}
The evidence  $\mathit{smokes}(\mathit{p3}) = \mathit{false}$ makes the third rule for \verb$smokes(p1)$ 
inactive. This in turn makes the probabilistic fact for \verb$influences(p3,p1)$ irrelevant. Nevertheless, the rules for \verb$smokes(p3)$ have to be in the
grounding, as the truth value of the head of a rule constrains the truth values of the bodies.
\end{example}

\subsection{The Boolean Formula for the Ground Program}
\label{sec:lp2bool}

We now discuss how to convert the rules in the relevant ground program $L_g$ to an equivalent Boolean formula $\varphi_r$. 
Converting a set of logic programming (LP) rules to an equivalent Boolean formula is a purely logical (non-probabilistic) problem. This has been well studied in the LP literature, where several conversions have been proposed, e.g.\ \citeN{Janhunen04}. Note that the conversion is not merely a syntactical rewriting issue; the point is that the rules and the formula are to be interpreted according to a different semantics. 
Hence the conversion should compensate for this: the rules under LP semantics (with Closed World Assumption) should be equivalent to the formula under FOL semantics (without CWA).

For \textbf{acyclic rules}, the conversion is straightforward, we can simply
take \emph{Clark's completion} of the rules \cite{Lloyd87,Janhunen04}. We illustrate this on the Alarm example, which is indeed acyclic.

\begin{example}[Formula for the {\tt alarm} rules]
\label{example:alarm:formula}
As shown in Example~\ref{example:alarm:steps}, the grounding of the Alarm example contains two rules for \verb$alarm$, namely \verb$alarm :- burglary$ and \verb$alarm :- earthquake$. Clark's completion of these rules is the
 propositional formula $alarm \leftrightarrow \mathit{burglary} \lor \mathit{earthquake}$, i.e., the alarm goes off if and only if there is burglary or earthquake. Once we
 have the formula, we often need to rewrite it in CNF form, which is straightforward for a completion formula. For the completion of \verb$alarm$, the resulting CNF has three clauses: $alarm \lor \lnot \mathit{burglary}$, $alarm \lor \lnot \mathit{earthquake}$, and $\lnot
alarm \lor \mathit{burglary} \lor \mathit{earthquake}$. The last clause reflects the CWA.\footnote{The Alarm example models a Bayesian network for the MARG task. For Bayesian networks, the problem of conversion to a weighted CNF formula has been considered before, and several encodings exist \cite{DarwicheBook,Sang05}. For ProbLog programs modelling Boolean Bayesian networks, like Alarm, our CNF encoding coincides with that of~\citeN{Sang05}.}
\end{example}

For \textbf{cyclic rules}, the conversion is more complicated. This
holds in particular for rules with \emph{positive loops}, i.e., loops with atoms that depend positively on each other, as in the recursive rule for \verb$smokes/1$. 
It is well-known that in the presence of positive loops, Clark's completion is not correct, i.e.\ the resulting formula is not equivalent to the rules \cite{Janhunen04}. 
\begin{example}[Simplified Smokers example]
\label{example:smokerssmall:ProbLog}
Let us focus on the Smokers program of Example~\ref{example:smokers:ProbLog}, but restricted to person p1 and p2.
\begin{verbatim}
0.2::stress(p1).             0.3::influences(p2,p1). 
0.2::stress(p2).             0.3::influences(p1,p2).                 
smokes(p1) :- stress(p1).
smokes(p1) :- smokes(p2), influences(p2,p1).
smokes(p2) :- stress(p2).
smokes(p2) :- smokes(p1), influences(p1,p2).
\end{verbatim}
Clark's completion of the rules for \verb$smokes(p1)$ and  \verb$smokes(p2)$ would
result in a formula which has as a model 
$\{ \mathit{smokes}(\mathit{p1}), \mathit{smokes}(\mathit{p2}), \lnot stress(\mathit{p1}), \lnot stress(\mathit{p2}),$ $\mathit{influences}(\mathit{p1},\mathit{p2}), \mathit{influences}(\mathit{p2},\mathit{p1}), \ldots \}$,
but this is not a model of the ground ProbLog program: the only model resulting from the total choice $\{ \lnot stress(\mathit{p1}), $ $ \lnot stress(\mathit{p2}),$ $\mathit{influences}(\mathit{p1},\mathit{p2}), \mathit{influences}(\mathit{p2},\mathit{p1}), \ldots \}$, is the model in which $\mathit{smokes}(\mathit{p1})$ and $\mathit{smokes}(\mathit{p2})$ are both false.
\end{example}

Since Clark's completion is inapplicable with positive loops, a range of more sophisticated conversion algorithms
have been developed in the LP literature. Since the problem is of a highly technical nature, we are unable to repeat
the full details in this paper. Instead, we briefly discuss the two conversion methods that we use in our work and refer to the corresponding literature for more details. 

Both conversion algorithms take a set of rules and construct an equivalent formula. The formulas generated by the two algorithms
are typically syntactically different because the algorithms introduce a
set of auxiliary atoms in the formula and these sets might differ. For
both algorithms, the size of the formula typically increases with the number of positive loops in the rules. The two algorithms are the following.
\begin{itemize}
\item The first algorithm is from the Answer Set Programming literature \cite{Janhunen04}. It first rewrites the given rules into an equivalent set of rules without positive loops (all resulting loops involve negation). This requires the introduction of auxiliary atoms and rules. Since the resulting rules are free of positive loops, they can be converted by taking Clark's completion. The result can then be written as a CNF. 
This algorithm is \textbf{rule based}, as opposed to the next algorithm.

\item The second algorithm was introduced in the LP literature \cite{Mantadelis10} and is \textbf{proof-based}. It first constructs all proofs of all atoms of interest, in our case all atoms in $\mathbf{Q} \cup \mathbf{E}$, using tabled SLD resolution. 
The proofs are collected in a recursive structure, namely a set of nested tries \cite{Mantadelis10}, which will have loops if the given rules had loops. The algorithm then operates on this structure in order to `break' the loops and obtain an equivalent Boolean formula. This formula can then be written as a CNF.
\end{itemize}

Both the rule-based and the proof-based conversion algorithm return a formula that is `equivalent' to the rules in $L_g$, in the sense of the following lemma.
\begin{lemma} \label{lemma:lp2bool}
Let $L_g$ be a ground ProbLog program. Let $\varphi_r$ denote the formula derived from the rules in $L_g$. Then $SAT(\varphi_r) = MOD(L_g)$.
\end{lemma}
Recall that  $MOD(L_g)$ denotes the set of models of a ProbLog program $L_g$, as defined in Section~\ref{sec:semantics}. On the formula side, we use $SAT(\varphi_r)$ to denote the set of models of a formula $\varphi_r$.\footnote{Both conversions for cyclic rules introduce additional or `auxiliary' atoms into $\varphi_r$. We can safely omit these atoms from the models in $SAT(\varphi_r)$ because both conversions are `faithful', so the truth value of auxiliary atoms is uniquely defined by the truth value of the original atoms. This means that the introduction of the auxiliary atoms does not create extra models. Hence, w.r.t.\ the original atoms we have the stated equivalence: $SAT(\varphi_r) = MOD(L_g)$. W.r.t.\ all atoms, $\varphi_r$ and $L_g$ are equisatisfiable.}


\begin{example}[Boolean formula for the simplified Smokers example]
Consider the ground program for the simplified Smokers example, given in Example~\ref{example:smokerssmall:ProbLog}. The proof-based conversion algorithm converts the ground rules in this program to an equivalent formula (in the sense of Lemma~\ref{lemma:lp2bool}) consisting of the conjunction of the following four subformulas.
\begin{eqnarray*}
\mathit{smokes}(\mathit{p1}) & \leftrightarrow & \mathit{aux1} \lor stress(\mathit{p1}) \\
\mathit{smokes}(\mathit{p2}) & \leftrightarrow & \mathit{aux2} \lor stress(\mathit{p2}) \\
\mathit{aux1} & \leftrightarrow & \mathit{smokes}(\mathit{p2}) \land \mathit{influences}(\mathit{p2},\mathit{p1})  \\
\mathit{aux2} & \leftrightarrow & stress(\mathit{p1}) \land \mathit{influences}(\mathit{p1},\mathit{p2})
\end{eqnarray*}
Here $\mathit{aux1}$ and $\mathit{aux2}$ are auxiliary atoms that are introduced by the conversion (though they could be avoided in this case). Intuitively, $\mathit{aux1}$ says that person $p1$ started smoking because he is influenced by person $p2$, who smokes himself. Note that while the ground program (in Example~\ref{example:smokerssmall:ProbLog}) is cyclic, the loop has been broken by the conversion process; this surfaces in the fact that the last subformula uses $stress(\mathit{p1})$ instead of $\mathit{smokes}(\mathit{p1})$.
\end{example}





\subsection{The Weighted Boolean formula}
\label{sec:wbool}

The final step of the conversion constructs the weighted Boolean formula starting from the Boolean formula for the rules $\varphi_r$. First, the formula $\varphi$ is defined as the conjunction of
$ \varphi_r$ and a formula $\varphi_e$ capturing the evidence
$\mathbf{E=e}$. Here $\varphi_e$ is a conjunction of unit clauses: there is a unit clause $a$ for each true atom and a clause $\lnot a$
for each false atom in the evidence. Second, we  define the weight
function for all literals in the resulting formula. The weight of a \emph{probabilistic literal} is derived from the probabilistic facts in the program: if the relevant ground program contains a probabilistic fact \verb$p::f$, then we assign weight $p$
to $f$ and weight $1-p$ to $\lnot f$. The weight of a \emph{derived literal} (a literal not
occuring in a probabilistic fact) is always $1$. 
The \emph{weight of a world $\omega$}, denoted $w(\omega)$, is defined to be the product of the weight of all literals in $\omega$.

\begin{example}[Weighted formula for Alarm]
\label{example:alarm:weight}
We have seen the formula for the Alarm program in Example~\ref{example:alarm:formula}. 
If we have evidence that $\mathit{calls}(\mathit{john})$ is true, we add a positive unit clause $\mathit{calls}(\mathit{john})$ to this formula (after doing this, we can potentially apply unit propagation to simplify the formula). Then we define the weight function. The formula contains three probabilistic atoms \verb$burglary$,  \verb$earthquake$ and
\verb$hears_alarm(john)$. The other atoms in the formula, \verb$alarm$ and \verb$calls(john)$, are derived atoms. Hence the weight function is as follows.\\
\begin{tabular}{ll}
$\mathit{burglary} \mapsto$ 0.1 & $\lnot \mathit{burglary} \mapsto$ 0.9 \\
$\mathit{earthquake} \mapsto$ 0.2 & $\lnot \mathit{earthquake} \mapsto$ 0.8 \\
$\mathit{hears\_alarm}(\mathit{john})  \mapsto$ 0.7 & $\lnot \mathit{hears\_alarm}(\mathit{john}) \mapsto$ 0.3 \\
$alarm \mapsto$ 1 & $\lnot \mathit{alarm} \mapsto$ 1 \\
$\mathit{calls}(\mathit{john})  \mapsto 1$ & $\lnot \mathit{calls}(\mathit{john})  \mapsto$ 1 \\
\end{tabular}
\end{example}

We have now seen how to construct the entire weighted formula from the relevant ground program. 
The following theorem states that this weighted formula is equivalent - in a particular sense - to the relevant ground program.  We will make use of this result  when performing inference on the weighted formula. 
\begin{theorem} \label{theorem:wbool}
Let $L_g$ be the relevant ground program for some ProbLog program with respect to $\mathbf{Q}$ and \mbox{$\mathbf{E}=\mathbf{e}$}. Let $MOD_{\mathbf{E=e}}(L_g)$ be those models in $MOD(L_g)$ that are consistent with the evidence $\mathbf{E=e}$. Let $\varphi$ denote the formula and $w(.)$ the weight function of the weighted formula derived from $L_g$. Then:
\begin{itemize}
\item[-] \textbf{(model equivalence)} $SAT(\varphi) = MOD_{\mathbf{E=e}}(L_g)$, 
\item[-] \textbf{(weight equivalence)} $\forall \omega \in SAT(\varphi)$: $w(\omega) = P_{L_g}(\omega)$, i.e., the weight of $\omega$ according to $w(.)$ is equal to the probability of $\omega$ according to $L_g$.
\end{itemize}
\end{theorem}
Note the relationship with Lemma~\ref{lemma:lp2bool} (p.~\pageref{lemma:lp2bool}): Lemma~\ref{lemma:lp2bool} applies to the formula $\varphi_r$ \emph{prior}
to asserting the evidence, whereas Theorem~\ref{theorem:wbool}
applies to the formula $\varphi$ \emph{after} asserting evidence.

\begin{example}[Equivalence of weighted formula and ground program]
The ground Alarm program of Example~\ref{example:alarm:steps} has three probabilistic facts and hence $2^3=8$ total choices and corresponding possible worlds. Three of these possible worlds are consistent with the evidence $\mathit{calls}(\mathit{john})=\mathit{true}$, namely the worlds resulting from choices in which $\mathit{hears\_alarm}(\mathit{john})$ is always true and at least one of $\{ \mathit{burglary}, \mathit{earthquake}\}$ is true. The reader can verify that the Boolean formula constructed in Example~\ref{example:alarm:weight} has exactly the same three models, and that weight equivalence holds for each of these models.
\end{example}

There is also a link between the weighted formula and \emph{Markov Logic Networks (MLNs)}. Readers unfamiliar with MLNs can consult \ref{app:MLN}. 
The weighted formula that we construct can be regarded as a ground MLN. The MLN contains the Boolean formula as a `hard' formula (with infinite weight). The MLN also has two weighted unit clauses per probabilistic atom: for a probabilistic atom $a$ and weight function $\{a \mapsto p, \lnot a \mapsto 1-p\}$, the MLN contains a unit clause $a$ with weight $\ln(p)$ and a unit clause $\lnot a$ with weight $\ln(1-p)$.\footnote{The values of the logarithms (and hence the weights) are negative, but any MLN with negative weights can be rewritten into an equivalent MLN with only positive weights \cite{MLN_PILPbook}.} 

\begin{example} [MLN for the Alarm example]
The Boolean formula $\varphi$ for our `Alarm' running example was shown in Example~\ref{example:alarm:steps}. 
The corresponding MLN contains this formula as a hard formula. The MLN also contains the following six weighted unit clauses. \\
\begin{tabular}{p{5cm}p{5cm}}
$\ln(0$.$1) \ \mathit{burglary}$ & $\ln(0$.$9) \ \lnot \mathit{burglary}$ \\
$\ln(0$.$2) \ \mathit{earthquake}$ & $\ln(0$.$8) \ \lnot \mathit{earthquake}$ \\
$\ln(0$.$7) \ \mathit{hears\_alarm}(\mathit{john})$ & $\ln(0$.$3) \ \lnot \mathit{hears\_alarm}(\mathit{john})$ \\
\end{tabular}
\end{example}

We have the following equivalence result.
\begin{theorem} \label{theorem:equiv_MLN}
Let $L_g$ be the relevant ground program for some ProbLog program with respect to $\mathbf{Q}$ and $\mathbf{E}=\mathbf{e}$. Let $\mathcal{M}$ be the corresponding ground MLN. The distribution $P(\mathbf{Q})$ according to $\mathcal{M}$ is the same as the distribution $P(\mathbf{Q} \mid \mathbf{E}=\mathbf{e})$ according to $L_g$.
\end{theorem}
Note that for the MLN we consider the distribution $P(\mathbf{Q})$ (not 
conditioned on the evidence). This is because the evidence is already hard-coded in the MLN.


\section{Inference on the Weighted Formula}
\label{sec:inference_section}

To solve the given inference task for the probabilistic logic program $L$, the query $\mathbf{Q}$ and evidence $ \mathbf{E}=\mathbf{e}$, we have converted the program to a weighted Boolean formula. 
A key advantage 
is that the inference task (be it MARG, MPE or EVID) can now be reformulated
in terms of well-known tasks such as weighted model counting or weighted MAX-SAT on the weighted formula. This implies that we can use any of the existing state-of-the-art algorithms for solving these tasks. In other words, by the conversion of ProbLog to weighted formula, we get the inference algorithms for free.  

\subsection{Task 1: Computing the probability of evidence (EVID)}

Computing the probability of evidence reduces to \emph{weighted model counting (WMC)}, a well-studied task in the SAT community. Model counting for a propositional formula is the task of computing the number of models of the formula. WMC is the generalization where every model has a weight and the task is to compute the sum of weights of all models. The fact that computing the probability of evidence $P(\mathbf{E}=\mathbf{e})$ reduces to WMC on our weighted formula can be seen as follows.
\[P(\mathbf{E}=\mathbf{e}) = \sum_{\omega \in MOD_{\mathbf{E}=\mathbf{e}}(L)} P_{L}(\omega) = \sum_{\omega \in SAT(\varphi)} w(\omega)\]
The first equality holds because $P(\mathbf{E}=\mathbf{e})$ by definition equals the total probability of all worlds consistent with the evidence. The second equality follows from Theorem~\ref{theorem:wbool}: model equivalence implies that the sets over which the sums range are equal, weight equivalence implies that the summed terms are equal. Computing $\sum_{\omega \in SAT(\varphi)} w(\omega)$ is exactly what WMC on the weighted formula $\varphi$ does. It is well-known that inference with Bayesian networks can be solved using WMC \cite{Sang05}. In \cite{FierensUAI11} we were the first to point out that this also holds for inference with probabilistic logic programs.
As we will see in the experiments, this approach improves upon state-of-the-art methods in probabilistic logic programming.

The above leaves open \emph{how} we solve the WMC problem. There exist many approaches to WMC, both exact \cite{DarwicheECAI04} and approximate \cite{SampleCount}. An approach that is particularly useful in our context is that of \emph{knowledge compilation}, `compiling' the weighted formula into a more `efficient' form.
While knowledge compilation has been studied for many different tasks \cite{DarwicheMarquisJAIR2002}, we need a form that allows for efficient WMC. Concretely, we compile the weighted formula into a so-called \emph{arithmetic circuit} \cite{DarwicheBook}, which is closely linked to the concept of  \emph{deterministic, decomposable negation normal form (d-DNNF)} \cite{DarwicheECAI04}. 

\subsubsection{Compilation to an Arithmetic Circuit via d-DNNF}
\label{sec:ac_d-DNNF}

We now introduce the necessary background on knowledge compilation and illustrate the approach with an example.

Knowledge compilation is concerned with compiling a logical formula, for which a certain family of inference tasks is hard to compute, into a representation where the same tasks are tractable (so the complexity of the problem is shifted to the compilation phase). In this case, the hard task is to compute weighted model counts (which is \#P-complete in general). After compiling a logical formula into a \emph{deterministic, decomposable negation normal form circuit (d-DNNF)} representation \cite{DarwicheECAI04} and converting the d-DNNF into an arithmetic circuit, the weighted model count of the formula can efficiently be computed, conditioned on any set of evidence. This allows us to compile a single d-DNNF circuit and evaluate all marginals efficiently using this circuit.

A negation normal form formula (NNF) is a rooted directed acyclic graph
in which each leaf node is labeled with a literal 
and each internal node is labeled with a conjunction 
or disjunction. 
A decomposable negation normal form (DNNF) is a NNF satisfying \emph{decomposability}: for every conjunction node, it should hold that no two children of the node share any atom with each other. 
A deterministic DNNF (d-DNNF) is a DNNF satisfying \emph{determinism}: for every disjunction node, 
all children\label{sec:inference} should represent formulas that are logically inconsistent with each other. For WMC, we need a d-DNNF that also satisfies \emph{smoothness}: for every disjunction node, all children should use exactly the same set of atoms. Compiling a Boolean formula to a (smooth) d-DNNF is a well-studied problem, and several compilers are available \cite{DarwicheECAI04,d_sharp_compiler}. 
These circuits are the most compact circuit language we know of today that supports tractable WMC~\cite{DarwicheMarquisJAIR2002}.

A d-DNNF is a purely logical construct. It is constructed by compiling the formula, irrespective of the associated weighting function. Hence a d-DNNF allows for model counting, but not for WMC. In order to do WMC, we need to convert the d-DNNF into an \emph{arithmetic circuit}, by taking into account the weighting function of our weighted formula. This conversion is done in two steps \cite{DarwicheBook}: 1)~replace all conjunctions in the internal nodes by multiplications, and all disjunctions by summations, 2)~replace every leaf node involving a literal $l$ by a subtree consisting of a multiplication node having two children, namely a leaf node with an \emph{indicator variable} for the literal $l$ and a leaf node with the weight of $l$ according the weighted formula. We now illustrate this for the Alarm example. 

\begin{figure*}
\subfigure[d-DNNF]{\includegraphics[scale=0.4]{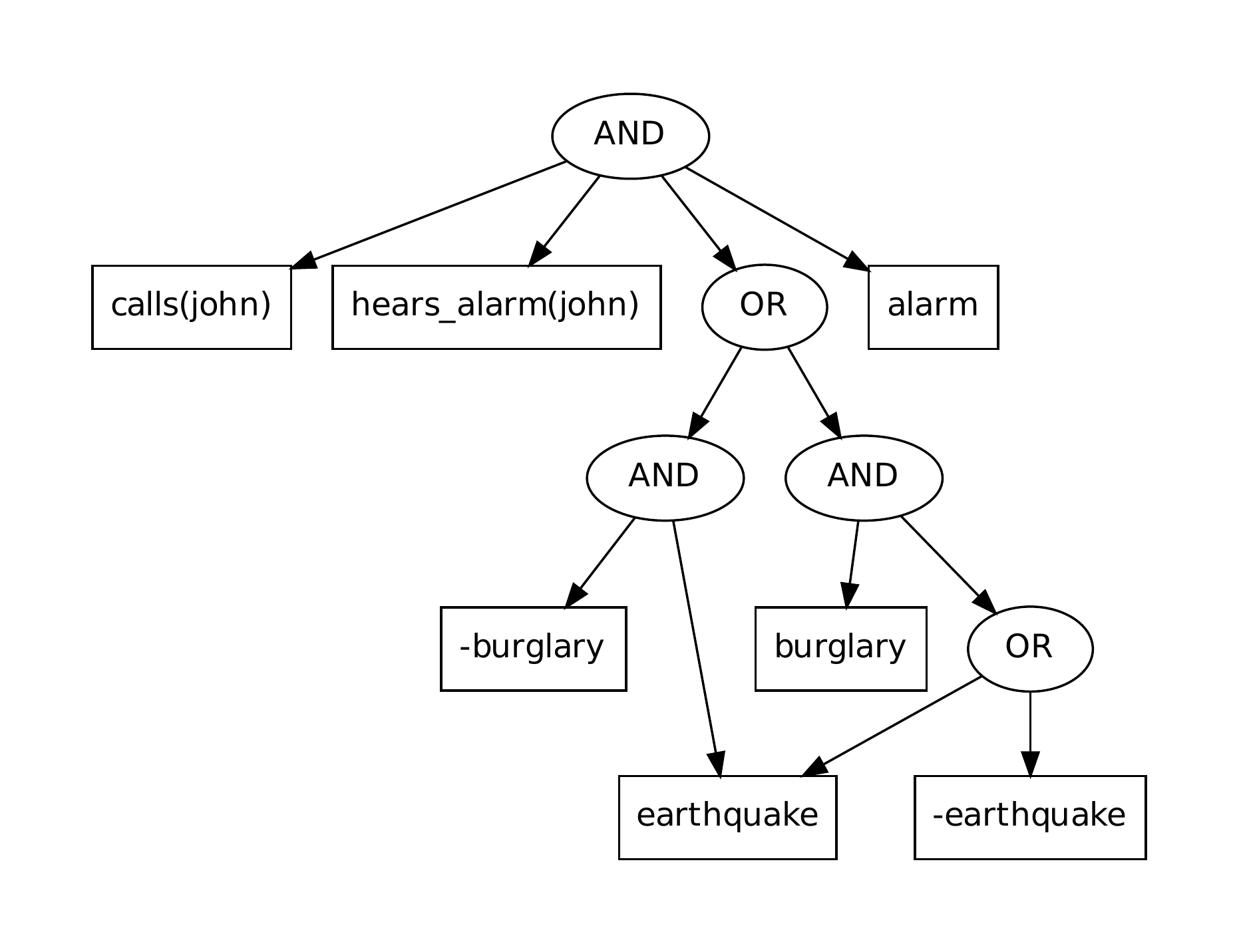}}
\subfigure[arithmetic circuit]{\includegraphics[scale=0.45]{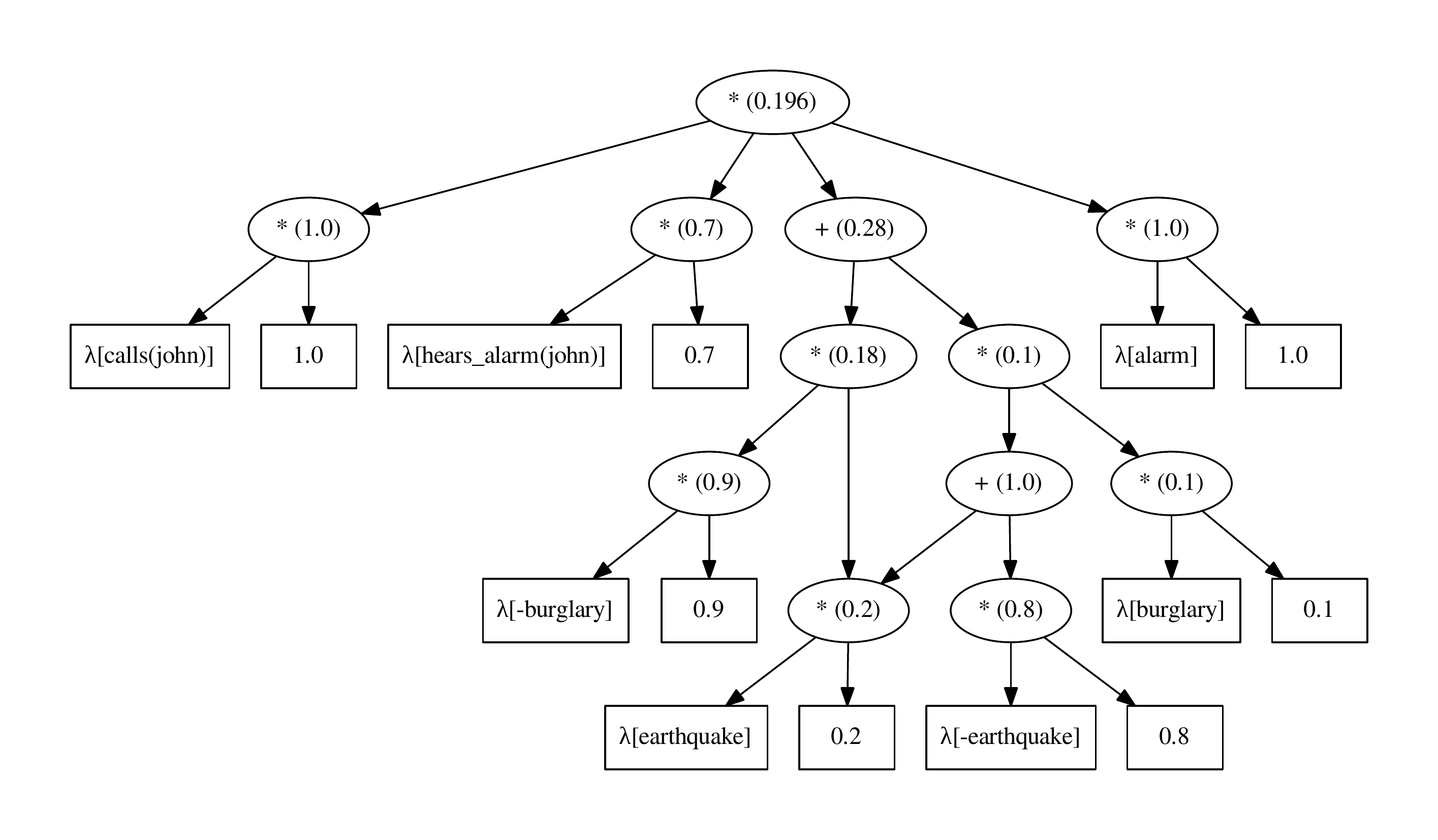}}
\caption{The d-DNNF for the Alarm example and the corresponding arithmetic circuit.} \label{fig:example5_ddnnf}
\end{figure*}

\begin{example}[d-DNNF and Arithmetic Circuit for the Alarm example]
\label{example:alarm:circuits}
We continue the Alarm example (Example~\ref{example:alarm:weight}). The formula for this example, under the evidence $\mathit{calls}(\mathit{john})=\mathit{true}$, is the conjunction of the following three subformulas.
\begin{eqnarray*}
&& alarm \leftrightarrow \mathit{burglary} \lor \mathit{earthquake} \\
&& \mathit{calls}(\mathit{john}) \leftrightarrow \mathit{alarm}, \mathit{hears\_alarm}(\mathit{john}) \\
&& \mathit{calls}(\mathit{john})
\end{eqnarray*} 
A corresponding d-DNNF is shown in Figure~\ref{fig:example5_ddnnf}a. Note that the AND-nodes in the d-DNNF (like the root note) indeed satisfy the property of decomposability; while the OR-nodes satisfy determinism. The function of the OR-node on the lower-right is to make the d-DNNF smooth.

The arithmetic circuit corresponding to this d-DNNF is shown in Figure~\ref{fig:example5_ddnnf}b. The values in brackets in the interal nodes will be used later and can be ignored for now. The $\lambda$-variables in the leaves are the indicator variables for the literals. The indicator variable for a literal $l$ is multiplied with a number, which is the weight of $l$ according to our weighting function. 
\end{example}

Now that we have an arithmetic circuit for our weighted formula, we are ready to perform WMC and compute the weighted model count $\sum_{\omega \in SAT(\varphi)} w(\omega)$. This count is found by simply \emph{evaluating} the arithmetic circuit: we instantiate all indicator variables to the value 1 and then bottom-up evaluate all nodes, until we arrive at the root node. The value found at the root is the desired weighted model count and also equals the probability of the evidence $P(\mathbf{E}=\mathbf{e})$.

\begin{example}[Evaluating the arithmetic circuit for the Alarm example]
We use the arithmetic circuit for the Alarm program given in Example~\ref{example:alarm:circuits}. Recall that this program and circuit were obtained using $\mathit{calls}(\mathit{john})=\mathit{true}$ as the evidence, so we can use this circuit to calculate the probability of evidence $P(\mathit{calls}(\mathit{john})$ $=\mathit{true})$. This is done by instantiating all indicator variables $\lambda$ to 1, and then evaluting the circuit. Figure~\ref{fig:example5_ddnnf}b illustrates this: the obtained values in each node are given between brackets. The value for the root is 0.196. This is the probability of evidence.
\end{example}

The above does not explain why we really need the indicator variables. The indicator variables allow us to add further evidence, on top of $\textbf{E}=\textbf{e}$, which is useful for MARG inference as we will see later. For instance, we can compute $P(\textbf{E}=\textbf{e} \land X = \mathit{true})$, for some additional atom $X$ in the arithmetic circuit, by setting the indicator variable $\lambda[X]$ to 1 and $\lambda[-X]$ to 0 when evaluating the circuit.\footnote{In a purely logical context, setting indicator variables to 0 corresponds to \textit{conditioning} the d-DNNF circuit.} 

\begin{figure}[htb]
    \centering
     \includegraphics[width=1\textwidth]{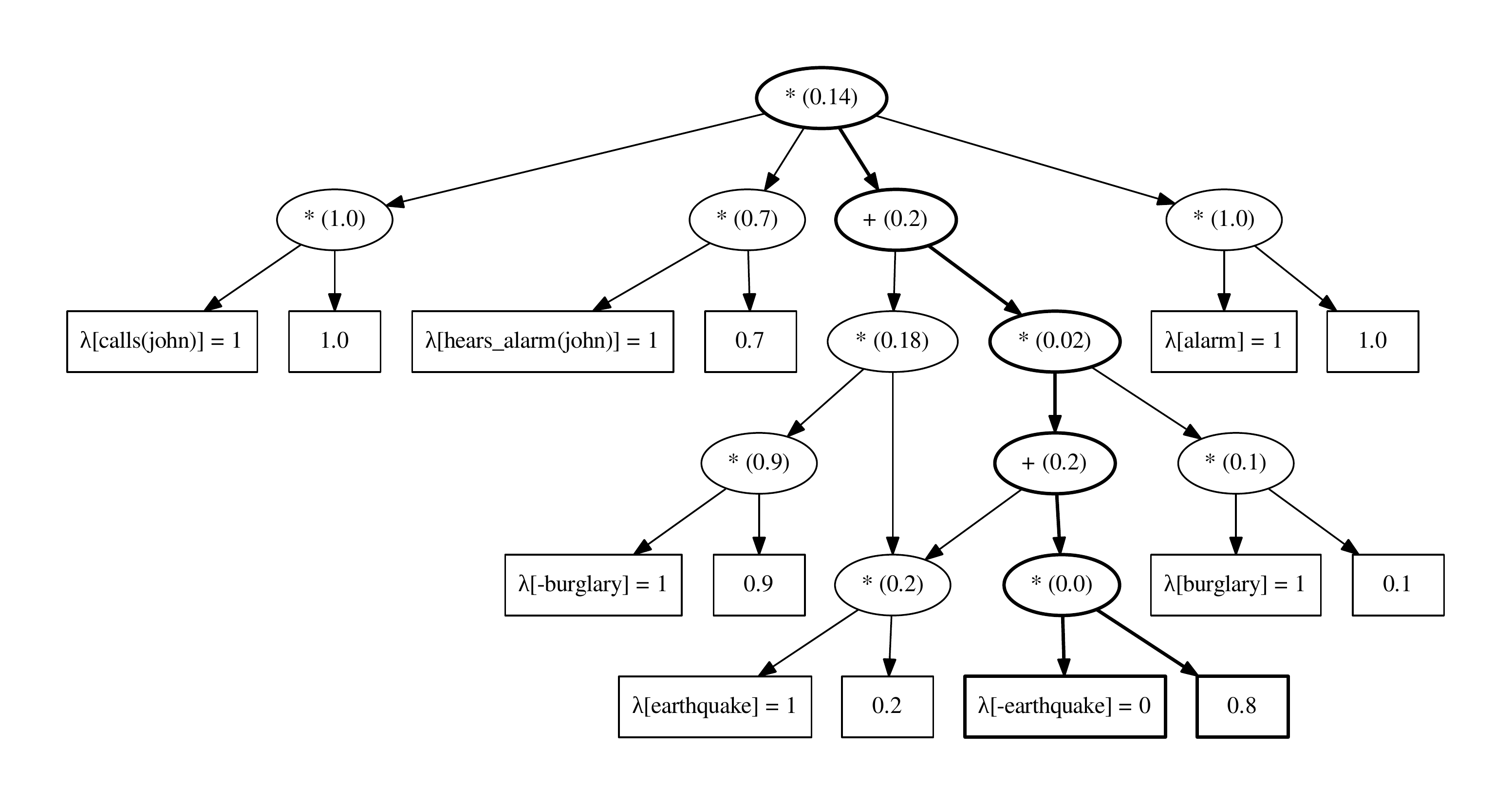}
     \caption{Evaluating an arithmetic circuit with additional evidence (the nodes which get a different value than in Figure~\ref{fig:example5_ddnnf}(b) are highlighted in boldface).}
    \label{fig:ACeval}
  \end{figure}

\begin{example}[Evaluating the arithmetic circuit in case of additional evidence]
Assume we want to compute $P(\mathit{calls}(\mathit{john}) =\mathit{true} \land \mathit{earthquake} = \mathit{true})$, using the same arithmetic circuit seen before, namely the circuit for $\mathit{calls}(\mathit{john}) =\mathit{true}$. Since we additionally have $\mathit{earthquake} = \mathit{true}$, we set $\lambda[\mathit{earthquake}]$ to 1, $\lambda[\mathit{-earthquake}]$ to 0, and all other indicator variables to 1 as before. The evaluation is illustrated in Figure~\ref{fig:ACeval}, yielding the result 0.14. Hence $P(\mathit{calls}(\mathit{john}) =\mathit{true} \land \mathit{earthquake} = \mathit{true}) =$ 0.14.
\end{example}

In the same way, the probability of any set of evidence can be computed, provided that this set extends the initial set $\textbf{E}=\textbf{e}$ (and that the additional atoms also appear in the compiled circuit). This also means that Step~3 of our conversion algorithm (Section~\ref{sec:wbool}), where we add the evidence $\varphi_e$ to the weighted Boolean formula, is not strictly needed: we can achieve the same result by using only the formula $\varphi_r$ (capturing the rules of the program) and setting the indicator variables in the circuit according to the evidence $\textbf{E}=\textbf{e}$. However, asserting the evidence $\varphi_e$ early makes the compilation phase more efficient (it allows for more unit propagation, etc).

    
In SRL, the work of Chavira et al.~\shortcite{chavira2006compiling} is closest to the approach in this section. They perform inference in \textit{relational Bayesian networks} by encoding them into a weighted Boolean formula and compiling this formula into an arithmetic circuit. The main difference is that relational Bayesian networks are not a programming language and assume acyclicity. That assumption greatly simplifies the step of converting to a weighted Boolean formula (cf.\ Section~\ref{sec:conversion}).

In summary, to compute the probability of evidence we 1) compile the formula to a d-DNNF, 2) convert the d-DNNF into an arithmetic circuit, 3) evaluate the arithmetic circuit.

\subsubsection{Compilation to an Arithmetic Circuit via BDD}

In the probabilistic logic programming (PLP) community, the state-of-the-art \cite{ProbLogIJCAI07} is to compile the program into another form, namely a \emph{reduced ordered Binary Decision Diagram (BDD)} \cite{Bryant86}. This approach is a special case of our above WMC approach (although it is usually not formulated like that; in fact, in \citeN{FierensUAI11} we were the first to point out the connection of the PLP-BDD approach to WMC).

A BDD is a special kind of d-DNNF, namely one that satisfies the additional properties of \emph{ordering} and \emph{decision}, see \citeN{DarwicheECAI04}. 
In our approach, we can alternatively replace the d-DNNF compiler by a BDD compiler. 
Computing the probability of evidence can then be done by either operating directly on the BDD, or by converting the BDD to an arithmetic circuit and evaluating the circuit (the first approach is merely a reformulation of the second). So while both compilation to BDD and d-DNNF are possible, there is theoretical and empirical evidence in the model counting literature that d-DNNFs outperform BDDs \cite{DarwicheECAI04}. Our experimental results confirm the superiority of d-DNNFs (Section~\ref{sec:experiments}).



We have now seen two ways of computing the probability of evidence: via d-DNNFs or BDDs. We will now see how this approach for computing the probability of evidence can be used as a building block for the MARG inference task (as is standard in the probabilistic literature).

\subsection{Task 2: Computing marginal probabilities (MARG)}
\label{sec:MARGsolution}

In MARG, we are given a set of query atoms $\mathbf{Q}$ and for each $Q \in \mathbf{Q}$ we need to compute $P(Q \mid \mathbf{E}=\mathbf{e})$. By definition $P(Q \mid \mathbf{E}=\mathbf{e}) = \frac{P(Q \land \mathbf{E}=\mathbf{e})}{P(\mathbf{E}=\mathbf{e})}$. Hence, if we have $N$ atoms in the query set $\mathbf{Q}$,  solving MARG reduces to computing the probability of the evidence, and computing $N$ probabilities of the form $P(Q \land \mathbf{E}=\mathbf{e})$, i.e., the probability of the conjunction of the evidence with a single atom. In the previous section, we have already seen how we can compute such probabilities from the compiled arithmetic circuit, by appropriately instantiating the indicator variables $\lambda$ and evaluating the circuit. The simplest approach is to apply this once for each query atom $Q \in \mathbf{Q}$ separately. However, we can solve this even more efficiently. Concretely, all required probabilities can be found \emph{in parallel}. To be precise, 
all probabilities of the form $P(X \land \mathbf{E}=\mathbf{e})$, with $X$ being any atom in the circuit, can be computed simultaneously by traversing the circuit twice (bottom-up and top-down). The required traversal algorithm can be found in the literature, see Algorithm 34 (simple version) and 35 (optimized version) in \citeN{DarwicheBook}. From this, we obtain all probabilities of the form $P(X \land \mathbf{E}=\mathbf{e})$. We then retain those that involve an atom from the query set ($X \in \mathbf{Q}$) and compute the required conditional probabilities $P(Q \mid \mathbf{E}=\mathbf{e})$ as $\frac{P(Q \land \mathbf{E}=\mathbf{e})}{P(\mathbf{E}=\mathbf{e})}$. As in the previous section, this entire approach can be performed using an arithmetic circuit derived from a compiled d-DNNF or from a BDD.

The knowledge compilation approach is typically used for exact inference. When dealing with large domains, we often need to resort to computing \emph{approximate} marginals. Approximate inference is often achieved by means of sampling techniques, such as Markov Chain Monte Carlo (MCMC). Standard MCMC approaches like Gibbs sampling cannot deal with weighted formulas because the formula itself is deterministic. Instead, we use the \emph{MC-SAT} algorithm that was developed specifically to deal with determinism \cite{MCSAT}. MC-SAT is an MCMC algorithm that in every step of the Markov chain calls a SAT solver to construct a new sample. MC-SAT takes an MLN as input. Theorem~\ref{theorem:equiv_MLN} ensures that if we apply MC-SAT on the appropriate MLN, we indeed obtain samples from the distribution $P(\mathbf{Q} \mid \mathbf{E}=\mathbf{e})$.

To summarize, we currently have three methods for the MARG task: exact inference by compilation to 1) d-DNNFs or 2) BDDs, or 3) approximate inference with MC-SAT.

\subsection{Task 3: Finding the most likely explanation (MPE)}

MPE is the task of finding the most likely interpretation (joint state) of all unobserved atoms given the evidence, i.e.\ finding $\argmax_{\mathbf{u}} P( \mathbf{U}=\mathbf{u} \mid \mathbf{E}=\mathbf{e})$, with $\mathbf{U}$ all unobserved atoms (i.e, all atoms in the ground program that are not in $\mathbf{E}$). MPE inference on weighted formulas has been studied before. We consider two approaches. 

The first approach is to perform MPE by means of \emph{knowledge compilation}. The compilation step (to compile an arithmetic circuit via a d-DNNF or BDD) is the same as before, only the traversal step differs.\footnote{For the MPE task, it is sufficient to compile into a DNNF circuit, which is not necessarily deterministic. DNNF circuits are potentially more succinct than d-DNNF circuits, but unfortunately there exist no compilers specifically for DNNF.} Again, the traversal algorithm can be found in the literature, see Algorithm 36 in \citeN{DarwicheBook}. This yields the exact MPE solution.\footnote{This approach yields the truth value of all ground atoms that occur in the relevant ground program (RGP) for the given evidence. All \emph{probabilistic atoms} that do not occur in the RGP are irrelevant w.r.t.\ the evidence (i.e., they are probabilistically independent of the evidence). Hence, for each of these atoms, we can simply independently chose the truth value with maximum probability according to the associated probabilistic fact. The truth value of all \emph{derived atoms} that do not occur in the RGP is then found by computing the well-founded model of the MPE total choice.}

The second approach is to perform MPE using techniques from the SAT solving community. Concretely, it is known that MPE 
reduces to \emph{partially weighted MAX-SAT} \cite{Park02}. A popular approximate approach for solving this task is stochastic local search \cite{Park02}. An example algorithm is \emph{MaxWalkSAT}, which is also the standard MPE algorithm for MLNs \cite{MLN_PILPbook}. 


Since our current ProbLog implementation focusses on MARG inference rather than MPE, we do not discuss these approaches in detail and will not consider them further in this paper.


\section{Learning Probabilistic Logic Programs from Partial Interpretations}
\label{sec:lfi}

We now present an algorithm for learning the parameters (the probabilities of the probabilistic facts) of a ProbLog program from data. We use the \emph{learning from interpretations (LFI)} setting.

\subsection{The Learning Setting}

Learning from (possibly partial) interpretations  is a common setting
in statistical relational learning, which has so far not yet been
studied in its full generality for probabilistic programming
languages (but see also \citeN{GutmannECML11}).

In the terminology used for inference 
in Section~\ref{sec:tasks}, partial interpretations
correspond to evidence, and hence, in this section we shall often use the term evidence instead of partial interpretation.
Let $\mathbf{At}$ be the Herbrand base, i.e., the set of all ground (probabilistic and derived) atoms in a given ProbLog program. In the fully observable case, we learn from a set of complete interpretations,
that is, the observed truth-values $\mathbf{e}$ of all the atoms in the Herbrand base $\mathbf{At}$ 
are given and the evidence variables $\mathbf{E}$ coincide with $\mathbf{At}$.
On the other hand, in  the partially observable case, we learn from a set of partial interpretations, that is,
we only observe the truth-values $\mathbf{e}$  of a set $\mathbf{E} \subset \mathbf{At}$ of observed atoms. 
We now develop an algorithm, called LFI-ProbLog, 
that learns from (possibly partial) interpretations of a ProbLog program. In a generative setting, one is typically interested in the maximum likelihood parameters given the training data. This can be formalized as follows.
\begin{description}
 \item[\textbf{Given:}]
 \item \begin{itemize}
   \item a ProbLog program  $T_{\mathbf{p}}$ containing a set of rules $R$ and a set of probabilistic facts $F = \{p_i::f_i\}$ with \emph{unknown} parameters $\mathbf{p} = \langle p_1, \ldots, p_N\rangle$
   \item a set of (possibly partial) interpretations $D=\{\mathbf{E_1} =\mathbf{e_1},\ldots, \mathbf{E_M}=\mathbf{e_M}\}$ (the training examples)
  \end{itemize}
 \item[\textbf{Find:}] the maximum likelihood probabilities  $\widehat{\mathbf{p}} = \langle\widehat{p_1}, \ldots ,\widehat{p_N}\rangle$, that is,
\begin{displaymath}
 \widehat{\mathbf{p}} = \arg\max\limits_{\mathbf{p}} P_{T_{\mathbf{p}}}(D) =\arg \max\limits_{\mathbf{p}} \prod_{m=1}^M P_{T _{\mathbf{p}}} (\mathbf{E_m} = \mathbf{e_m})
\end{displaymath}
where $ P_{T_{\mathbf{p}}}(\mathbf{E_m}=\mathbf{e_m})$ is the probability of evidence $\mathbf{E_m}=\mathbf{e_m}$ in the ProbLog program $T_{\mathbf{p}}$ with parameters $\mathbf{p}$.
\end{description}

Example~\ref{example:lfi_alarm} illustrates the LFI 
setting using the Alarm program from Example~\ref{example:alarm}.

\begin{example}[Learning From Interpretations]
\label{example:lfi_alarm}
\small
\begin{verbatim}
P1::burglary.                        person(mary).        alarm:- burglary.
P2::earthquake.                      person(john).        alarm:- earthquake.
P3::hears_alarm(X):- person(X).      calls(X) :- alarm, hears_alarm(X).
\end{verbatim}
\normalsize
A ProbLog program is given in which the probabilities \verb=P1=, \verb=P2= and \verb=P3= are unknown and should be learned from partial interpretations, which contain the truth value for some of the atoms: $\{alarm=true\},\{earthquake=true, calls(mary)=true\},\{calls(john)=true\}$. The goal is to find the probabilities \verb=P1=, \verb=P2= and \verb=P3= such that the combined probability of the partial interpretations is maximal.
\end{example} 

One has to consider two cases when computing the maximum likelihood parameters $\widehat{\mathbf{p}}$. In the fully observable case where the truth values for each of the atoms in the Herbrand base is known, one can obtain $\widehat{\mathbf{p}}$ by counting. In the more complex case of partial interpretations, one has to use an approach such as Expectation Maximization to deal with the partial observability.

\subsection{Full Observability}
\label{subsec:fullobs}
In the fully-observable case, the maximum likelihood estimate $\widehat{p_n}$ for a 
probabilistic fact 
$p_n::f_n$ can be calculated directly from the interpretations. 
When $p_n::f_n$ is intensional, it represents multiple ground instances, that is, probabilistic facts: $p_n::f_{n,1}, \ldots, p_n::f_{n,K_n^m}$ where $K_n^m$ is the number of ground instances represented by the fact $p_n::f_n$ in interpretation $\mathbf{E_m} = \mathbf{e_m}$. When $p_n::f_n$ is ground and extensional, $K_n^m$ is equal to 1 and the fact represents itself only. The maximal likelihood estimates can be calculated using the following formula.

\begin{eqnarray}
\label{eq:mstepfullobs}
\widehat{p_n}  =\frac{1}{Z_n} \sum\limits_{m=1}^M\sum\limits_{k=1}^{K_n^m}\delta_{n,k}^m & \textrm{where} &   \delta_{n,k}^m =
	  \left\{\begin{array}{ll}
           1 & \text{if }f_{n,k}=true \in \mathbf{E_m}=\mathbf{e_m}\\
           0 & \text{if }f_{n,k}=false \in \mathbf{E_m}=\mathbf{e_m}
         \end{array}\right.
\end{eqnarray}

The sum is normalized by $Z_n=\sum_{m=1}^M {K_n^m} $, the total number of probabilistic facts represented by $f_n$ in all training examples.  When 
$Z_n$ is 0 in the data, $\widehat{p_n}$ is not calculated (there is no data to estimate it from).

\subsection{Partial Observability}
\label{subsec:partialobs}
In many applications the training examples are only partially observed. In the alarm example, we may receive a phone call but we may not know whether an earthquake has occurred.  In the partially-observable case -- similar to Bayesian networks -- it is impossible to compute the maximum likelihood estimates in closed-form. 
Instead, we use the \emph{Expectation Maximization (EM)}, see Algorithm~\ref{alg:lfi:mainloop}. 
In this algorithm, the parameters $p_n^0$ are initialized randomly. During each iteration $i$, the ProbLog program $T_{\mathbf{p}^i}$ with parameters $\mathbf{p}^i$ is used to estimate the probability of the unobserved atoms being true in each interpretation, $P_{T_{\mathbf{p}^i}}(f_{n,k}|\mathbf{E_m}=\mathbf{e_m})$ (the expectation step). These expectations are then used as 
to update the parameters of the program using the following equation 
(the maximization step).

\begin{equation}
\label{eq:msteppartialobs}
p_n^{i+1}  = \frac{1}{Z_n} \sum\limits_{m=1}^M\sum\limits_{k=1}^{K_n^m} P_{T_{\mathbf{p}^i}}(f_{n,k}|\mathbf{E_m}=\mathbf{e_m})
\end{equation}

Algorithm~\ref{alg:lfi:mainloop} uses the inference mechanism described in Section \ref{sec:MARGsolution} for computing the marginals in the expectation step. We can make two optimizations. Firstly, for the facts $f_{n,k}$ that are not contained in the dependency set of a partial interpretation $\mathbf{E_m}=\mathbf{e_m}$, the probability $P_{T_{\mathbf{p}^i}}(f_{n,k}|\mathbf{E_m}=\mathbf{e_m})$ is equal to $p^i_n$. These facts slow down the updating process
and should therefore not be included in the sum. This can be realized by compiling the d-DNNF for the query $P_{T_{\mathbf{p}^i}}(\mathbf{E_m}=\mathbf{e_m})$ and to use the resulting d-DNNFs to compute the marginal probabilities $P_{T_{\mathbf{p}^i}}(f_{n,k}|\mathbf{E_m}=\mathbf{e_m})$ only for those facts $f_{n,k}$ included in the d-DNNF. For example, when we compile the d-DNNF for the third partial interpretation of Example~\ref{example:lfi_alarm}, we obtain the ground program from Example \ref{example:alarm:steps} and the d-DNNF from Figure~\ref{fig:example5_ddnnf}a. This d-DNNF does not contain \verb=calls(mary)= so this atom will not be used to update the probabilities for the third partial interpretation. When no groundings for a learnable fact are present in any of the d-DNNFs, a zero probability is learned as no information is given. Secondly, one can observe that changing the parameters of a ProbLog program does not change the structure of the compiled d-DNNFs. This means that the d-DNNFs that have been compiled in the first iteration can be reused in all further iterations. The algorithm keeps on updating the parameters until the log likelihood of the interpretations is maximal. Each iteration of the algorithm is guaranteed to improve the likelihood of the data.

\begin{algorithm} [h]
  \caption[LFI-ProbLog]{The main loop of LFI-ProbLog. The ProbLog program is compiled into a d-DNNF for each partial interpretation $\mathbf{E_m} = \mathbf{e_m}$. After the compilation step, the algorithm follows an EM update scheme, first using the current model to complete the data and then estimating the new model parameters from the resulting counts until convergence.}
  \label{alg:lfi:mainloop}
  \begin{algorithmic}[1]
    \Function{LFI-ProbLog}{$T = \{p_1::f_1, \ldots, p_N::f_N\} \cup R,D=\{\mathbf{E_1} =\mathbf{e_1},\ldots, \mathbf{E_M}=\mathbf{e_M}\}$}
    \For{$1 \le n \le N$}
    \State $p_n^0 \leftarrow rand(0,1)$ \Comment{The fact probabilities are initialized with a random probability}
    \EndFor
    \For{$1 \le m \le M$} \Comment{Loop over training examples}
    \State d-DNNF$_m$ $\leftarrow$ \Call{compile}{$P_{T_0}(\mathbf{E_m} = \mathbf{e_m})$}
    \EndFor
    \State $i \leftarrow 0$ \label{alg:lfi:mainloop:i}
    \While{not converged} \Comment{EM algorithm}
    \State$i \leftarrow i + 1$
    \For{$1 \le m \le M$}
    \For{$1 \le n \le N$} \label{alg:lfi:mainloop:n}
    \For{$1 \le k \le K_n^m$}  \label{alg:lfi:mainloop:k}
    \State compute $P_{T_{i-1}}(f_{n,k}|\mathbf{E_m})$ using d-DNNF$_m$ \Comment{E Step}  \label{alg:lfi:mainloop:bb}
    \EndFor
    \EndFor
    \EndFor
    \For{$1\le n \le N$} \Comment{Loop over probabilistic facts}
   \State $p_n^i \leftarrow \frac{1}{Z_n} \sum\nolimits_{m=1}^M\sum\nolimits_{k=1}^{K_n^m}P_{T_{i-1}}(f_{n,k}|\mathbf{E_m})$ \Comment{M Step (cf.~Eq.~\ref{eq:msteppartialobs})}
    \EndFor
   \EndWhile  \label{line:lfi:emstop}
   \State \Return $\{p^i_n ::f_n\ |\ f_n\in F \}\cup R$ 
    \EndFunction
  \end{algorithmic}
\end{algorithm}

The learning algorithm uses a black box for the MARG inference task (line~\ref{alg:lfi:mainloop:bb}). In principle, any inference algorithm will work, including approximate ones. However, by choosing knowledge compilation for inference, we need to compile a circuit only once for each training example. This is the hard task. Once we have a circuit, computing expectations becomes easy, and we can reuse the circuit many times, for all $i$, $k$ and $n$ in lines~\ref{alg:lfi:mainloop:i}, \ref{alg:lfi:mainloop:n} and~\ref{alg:lfi:mainloop:k} of Algorithm~\ref{alg:lfi:mainloop}.
Furthermore, all marginal probabilities $P_{T_{\mathbf{p}}}(f_{n,k}|\mathbf{E_m})$ for the same evidence set $\mathbf{E_m}$ and parameterization $\mathbf{p}$ can be computed at once, in only two passes over the d-DNNF circuit.

\subsection{Discussion}

The above learning algorithm is based on the LFI algorithm that we developed in earlier work \cite{GutmannECML11}, see the discussion in Section~\ref{sec:introduction}. Our new algorithm has some advantages over the earlier version. First, the new algorithm can deal with cyclic programs (the old algorithm uses Clark's completion, which applies only to acyclic programs, cf.\ Section~\ref{sec:lp2bool}). Second, the new algorithm scales better as it employs a more efficient approach for inference in the expectation step, namely compilation to d-DNNFs instead of to BDDs (see the experiments in Section~\ref{sec:exp_BDD_dDNNF}). Furthermore, the new description of the algorithm more clearly separates the learning from the inference steps.

The complexity of parameter learning (and of MARG and MPE inference) is worst-case exponential in the \emph{treewidth}~\cite{robertson1986graph} of the weighted Boolean formula when using knowledge compilation to d-DNNF~\cite{darwiche2001tractability}. This theoretical complexity bound is in line with the complexity of classical algorithms for inference and learning in probabilistic graphical models.
For example, hidden Markov models have a constant treewidth in terms of the number of time steps considered.
Learning the parameters of these models is linear in the number of time steps, both when using LFI-ProbLog with d-DNNF compilation and, for example, expectation maximization with the classical junction tree algorithm.
These bounds assume that both algorithms succeed at finding the optimal tree decomposition of the model, which itself is a hard task in theory. In practice, however, there exist heuristics that can find good tree decompositions of many different kinds of models.


\section{Implementation of the System ProbLog2}
\label{sec:implementation}


The first ProbLog system \cite{KimmigTPLP11} focused on the inference task of computing the success probability of a single atom (Section~\ref{sec:tasks}) and on learning from entailment \cite{GutmannKKR08}. ProbLog2, the new ProbLog system described in this paper, focusses on different tasks, namely computing marginal probabilities and the probability of evidence, as well as learning from interpretations. This new setting is closer in spirit to the work on graphical models and Statistical Relational Learning (like Markov Logic). As a consequence of this new emphasis, the design of ProbLog2 is quite different from that of the first ProbLog. The implementation of the first ProbLog was tightly integrated in YAP Prolog \cite{KimmigTPLP11}. In contrast, ProbLog2 consists of a number of relatively loosely-coupled components, and involves almost no Prolog code. This new design is closer in spirit to that of some Answer Set Programming systems than to Prolog. 

We now briefly discuss the different components of the implementation. Most of these components are existing state-of-the-art programs, rather than being tailor-made for ProbLog. 
\begin{itemize}
\item The \emph{grounding} component computes the relevant ground program from the given ProbLog program (and query and evidence). This is the only component that is written in (YAP) Prolog. It is essentially a meta-interpreter that collects proofs to construct the dependency set (Section~\ref{sec:RGP}).
\item The \emph{conversion} component converts the rules in the relevant ground program to a Boolean (CNF) formula. The user can choose between the proof-based and the rule-based conversion (Section~\ref{sec:lp2bool}). For the proof-based conversion \cite{Mantadelis10}, we use our own implementation. For the rule-based conversion, we use the code of \citeN{Janhunen04}, as used in the Answer Set Programming community. 
\item The \emph{exact inference} component is based on knowledge compilation and consists of two parts: a compiler and an evaluation algorithm. For compilation to d-DNNF, the user can choose between the `c2d' compiler by \citeN{DarwicheECAI04} or the more recent `DSHARP' compiler \cite{d_sharp_compiler}.\footnote{All experiments in this paper use c2d.} For compilation to a BDD, we use CUDD (see \mbox{\texttt{http://vlsi.colorado.edu/$\sim$fabio/CUDD/}}). For constructing and evaluating the corresponding arithmetic circuit we use our own code.
\item The \emph{approximate inference} component converts the weighted formula to a Markov Logic Network and then uses the MC-SAT \cite{MCSAT} code from the Alchemy package to perform the sampling.
\item The \emph{learning component}, LFI-ProbLog, builds heavily on the inference component, as explained before (Section~\ref{sec:lfi}). It is essentially an Expectation Maximization loop around the inference component.
\end{itemize}
As mentioned, the above components are relatively loosely-coupled. They are bundled into a pipeline by means of Python code. A major advantage of our design is that it allows to build an entire ProbLog system by (mostly) using existing state-of-the-art programs for the different components, such as Janhunen's conversion program and the various d-DNNF and BDD compilers. Moreover, research on conversion of logic programs, knowledge compilation, weighted model counting, etc, continues, with new tools being released. Whenever a new tool becomes available for a particular component, we can benefit from this, and integrate it into our system. Such a design of course also has drawbacks. The two main drawbacks are that there is a certain latency between the components because of I/O issues, and that the system is complex to install and configure because of the different components written in different programming languages. 

ProbLog2 is available on \mbox{\texttt{http://dtai.cs.kuleuven.be/problog/}}.\footnote{In addition to the MARG, MPE and learning tasks, the ProbLog2 system supports MAP and decision-theoretic inference~\cite{VdBAAAI10}, which are not discussed here.}



\section{Experiments}
\label{sec:experiments}

The goal of our experiments is to establish the feasibility of our approach, 
and to analyze the influence of the different parameters. We focus on MARG inference and learning. Concretely, we aim to answer six questions. 
\begin{itemize}
\item[{\bf Q1}] Is working with the relevant rather than the complete ground program more efficient? 
\item[{\bf Q2}] Which of the two considered algorithms for converting the ground program to a Boolean formula (rule-based or proof-based conversion) performs best?
\item[{\bf Q3}] Which of the two considered approaches for knowledge compilation (using d-DNNFs or BDDs) performs best?
\item[{\bf Q4}] When computing success probabilities (the `classical' ProbLog setting), does ProbLog2 
outperform the previous ProbLog implementation?
\item[{\bf Q5}] When learning from data generated from a known ProbLog program, can we recover the parameters of the original program given a reasonable amount of data?
\item[{\bf Q6}] When learning from real-world data, can we obtain results comparable to the ones obtained with a state-of-the-art system (namely Alchemy)?
\end{itemize}
Note that in {\bf Q6} we compare our system to Alchemy, which is the standard system for Markov Logic (see \mbox{\texttt{http://alchemy.cs.washington.edu/}}).

\subsection{Programs and Datasets}
\label{sec:exp:data}

We perform experiments on three types of applications.

\paragraph{Social networks.} We use the standard \emph{Smokers} application \cite{MLN_PILPbook}. The ProbLog program contains the following intensional probabilistic facts and rules.
\begin{verbatim}
0.2::stress(P) :- person(P).                           
0.3::influences(P1,P2) :- friend(P1,P2). 
0.1::cancer_spont(P) :- person(P).                           
0.3::cancer_smoke(P) :- person(P). 
         
smokes(X) :- stress(X).
smokes(X) :- smokes(Y), influences(Y,X).
cancer(P) :- cancer_spont(P).
cancer(P) :- smokes(P), cancer_smoke(P).
\end{verbatim}
In addition, the program contains ground (non-probabilistic) facts for the predicates $\mathit{person}/1$ and $\mathit{friend}/2$. The number of such facts is varied; see the next section.

\paragraph{Collective classification.} We use the relational WebKB dataset.\footnote{See \mbox{\texttt{http://www.cs.cmu.edu/$\sim$webkb/}}.} In WebKB, university web pages need to be tagged with classes (like course page, student page, etc). This is modeled with a predicate $\mathit{has\_class}(\mathit{Page},\mathit{Class})$. The rules in the ProbLog program are the following; they specify how the class $C$ of a page $P$ depends on the textual content of $P$ (the words $W$ on the page), and on the classes of pages that link to~$P$. 
\begin{verbatim}
has_class(P,C) :- word_class(P,W,C).
has_class(P,C) :- has_class(P2,C2), link_class(P,P2,C,C2).
\end{verbatim}
For the predicate $\mathit{word\_class}/3$, there is a different intensional probabilistic fact for every (word,class)-pair in the dataset. Each such intensional probabilistic fact looks as follows.
\begin{verbatim}
prob::word_class(P,word1,class1) :- has_word(P,word1).
\end{verbatim}
The reason why we need a different intensional probabilistic fact for each (word,class)-pair is that for every such pair the involved probability (\verb;prob;) can be different. Similarly, for the predicate $\mathit{link\_class}/4$, there is one intensional probabilistic fact for every pair of classes in the dataset.
\begin{verbatim}
prob::link_class(P,P2,class1,class2) :- links_to(P,P2).
\end{verbatim}
The predicates that occur in the `bodies' of these intensional probabilistic facts ($\mathit{has\_word}/2$ and $\mathit{links\_to}/2$) are defined in the dataset. The probabilities of the probabilistic facts were learned from data using LFI-ProbLog.

\paragraph{Probabilistic grids.} For comparing ProbLog2 to the previous ProbLog implementation, we use the classical ProbLog application of probabilistic graphs \cite{ProbLogIJCAI07}. The program represents a graph in which edges are labelled with a probability. Here we use a \emph{$16 \times 16$ grid} as the graph. This consists of nodes $n_{x,y}$, with $x,y \in \{1,\dots,16\}$, lined out on a square grid with horizontal, vertical and diagonal directed edges between adjacent nodes. Concretely, the edges are the following. 
\begin{eqnarray*}
n_{x,y} \rightarrow n_{x+1,y} \ \ \ & \forall x \in \{1,\dots,15\}, y \in \{1,\dots,16\} & (\mathit{horizontal}) \\
n_{x,y} \rightarrow n_{x,y+1} \ \ \ & \forall x \in \{1,\dots,16\}, y \in \{1,\dots,15\} & (\mathit{vertical}) \\
n_{x,y} \rightarrow n_{x+1,y+1} & \forall x \in \{1,\dots,15\}, y \in \{1,\dots,15\} & (\mathit{diagonal}) \\
\end{eqnarray*}
Every edge has probability 0.5. Such a probabilistic graph is modelled in ProbLog by a set of probabilistic $\mathit{edge/2}$ facts. For instance, the horizontal edge from $n_{1,1}$ to $n_{2,1}$ is represented  as the probabilistic fact \verb;0.5::edge(n_1_1,n_2_1);. The goal is to find the probability of there being a path between certain nodes in the graph, where path is defined in the usual Prolog way.
\begin{verbatim}
path(X,Y) :- edge(X,Y).
path(X,Y) :- edge(X,Z), path(Z,Y).
\end{verbatim}


\subsection{Experimental Setup}

We now describe how we use these three programs (Smokers, WebKB and grids)  in our experimental setup.

\subsubsection{Inference Setup}
\label{sec:inf_setup}

\paragraph{MARG inference.} We test MARG influence on Smokers and WebKB. The main parameter influencing the complexity of our inference
experiments is the `domain size', i.e., the number of constants considered. For Smokers, the domain size is the number of people; for WebKB, it is the number of webpages (we take subsets of all pages that occur in the dataset). In our experiments, we vary the domain size and see how different measures, such as runtime, evolve. For each considered domain size we generate multiple different instances of the MARG task (10 for Smokers, 8 for WebKB), as described below. We report median results over these different instances (we use median because it is more stable than arithmetic average).

Given a particular domain size, one instance of the MARG task is generated as follows. \textbf{(1)} For both Smokers and WebKB, the program involves ground (non-probabilistic) facts for certain `background' predicates. We first generate interpretations for these predicates. For Smokers, the background predicate is $\mathit{friend}/2$, which determines the actual social network. We use a generator of synthetic power law random graphs (since such graphs are known to resemble real social networks) and convert the obtained graph to $\mathit{friend}/2$ facts. For WebKB, the background predicates are $\mathit{has\_word}/2$ and $\mathit{links\_to}/2$, for which interpretations can be found in the dataset. \textbf{(2)} Given the domains and background facts, we select the set of query and evidence atoms, $\mathbf{Q}$ and $\mathbf{E}$. For Smokers, we use 50\% of all $\mathit{smokes}/1$ and $\mathit{cancer}/1$ atoms as evidence and the other $smokes/1$ and $cancer/1$ atoms as queries. All atoms for the other predicates are neither query nor evidence. For WebKB, we have a similar setup: we use 50\% of all $\mathit{has\_class}/2$ atoms as evidence and the other $\mathit{has\_class}/2$ atoms as query. \textbf{(3)} The previous step generates the sets $\mathbf{Q}$ and $\mathbf{E}$, but not yet the values for the evidence atoms, i.e. the vector of truth values $\mathbf{e}$. To do so, we generate a `sample' of the ProbLog program. This is done by independently sampling each ground probabilistic fact, and then computing the well-founded model of the resulting logic program (as dictated by the ProbLog semantics). The result is a complete interpretation of all predicates in the program. From this interpretation, we extract the truth values of all atoms in the evidence set $\mathbf{E}$, and we use these truth values to construct the vector $\mathbf{e}$. (We similarly store the values of all atoms in the query set $\mathbf{Q}$ because we need them later as `query ground truth'; see Section~\ref{sec:MCSATexp}).

\paragraph{Special case: success probability.} The above is for MARG inference in the presence of multiple queries and evidence. In addition we also perform an experiment in the classical \emph{success probability} setting, where the goal is to compute the probability of a single query, without evidence \cite{KimmigTPLP11}. For this experiment, we use the probabilistic grid program. Per experiment, we ask a single query of the form \verb;path(n_i_i,n_16_16); where $i$ is being varied from $1$ to $15$. In other words: we are asking for the probabibity of there being a path from a node $n_{i,i}$ on the diagonal of the grid to $n_{16,16}$, the lower right corner of the grid. The smaller the value of $i$, the longer these paths become (and the more possible paths there are), and hence the harder the computation. We measure the effect of the value of $i$ on the runtime of the query in both ProbLog2 and the first ProbLog implementation (\mbox{\texttt{http://dtai.cs.kuleuven.be/problog/}}). For each value of $i$, we repeat the experiment 10 times and average the measured runtimes.




\subsubsection{Learning Setup}

In the learning experiments, we estimate the probabilities of all probabilistic facts from data.

For Smokers, we again vary the domain size. For each size we generate 170 experiments. We sample 40, 50, \dots, 200 interpretations from which we retain 10, 40, 70 and 100 percent of the atoms together with their truth value in the partial interpretations. From these interpretations we learn the probabilities for all intensional probabilistic facts in the program (predicates \verb=stress/1=, \verb=influences/2=, \verb=cancer_spont/1= and \verb=cancer_smoke/1=). 

For WebKB, the dataset consists of four disjoint sets of webpages, one per university. Per university, we use only the 20 words that contain the most information (as measured by information gain with respect to the class labels). We perform four-fold cross validation using both the Alchemy system (with a standard MLN for this application) and LFI-ProbLog.

The ProbLog program that we use for learning is slightly different from the one we use for inference. In addition to the rules given earlier (Section~\ref{sec:exp:data}), we include in the program two more causes for a page to have a certain class.
\begin{verbatim}
has_class(P,C) :- fixed_prior(P,C).
has_class(P,C) :- learnable_prior(P,C).
\end{verbatim}
The predicate $\mathit{learnable\_prior}/2$ accounts for the pages that are tagged with a class that can not be explained through words and links. There is one such probabilistic fact for each class.
\begin{verbatim}
prob::learnable_prior(P,class1) :- page(P).
\end{verbatim}
The predicate $\mathit{fixed\_prior}/2$ makes sure that every page can be tagged with every class.
\begin{verbatim}
0.001::fixed_prior(P,C) :- page(P), class(C).
\end{verbatim}
Finally, for computational reasons, we modify the rule that spreads influence across links ($\mathit{link\_class}/4$) such that pages can only influence their direct neighbors.

We learn all \verb;prob;-parameters in the program (not the probability of $\mathit{fixed\_prior}/2$). The learned program is too big to perform exact inference. Hence, when evaluating the learned program (which requires running inference), we use an approximation, namely we remove all probabilistic facts with a learned probability below 0.05.

\subsection{Experimental Results}

We now discuss our results in terms of the six questions raised earlier.

\begin{figure*}
\subfigure[runtime of conversion to a weighted Boolean formula]{\includegraphics[height=8.0cm,width=3.4cm,angle=270]{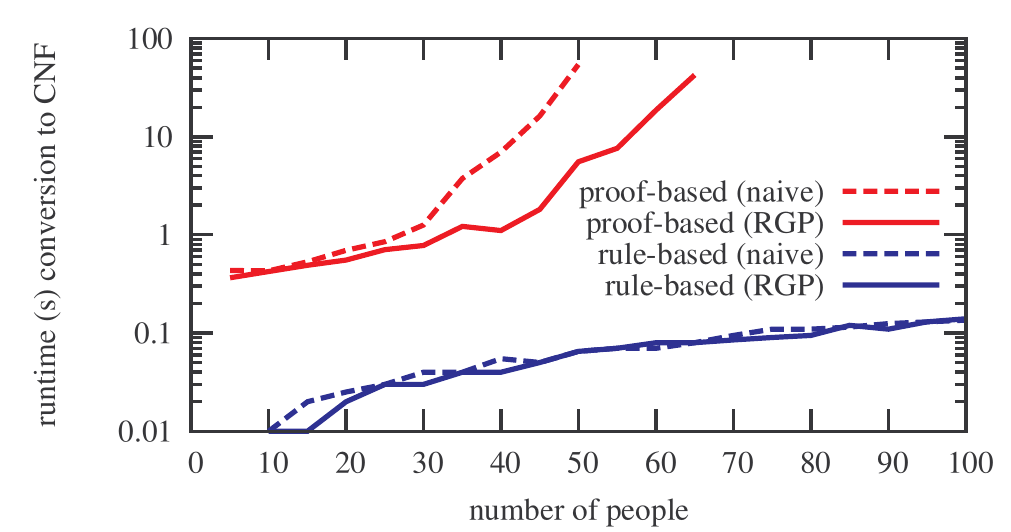}} 
\subfigure[size of the Boolean formula (number of clauses in the CNF)]{\includegraphics[height=8.0cm,width=3.4cm,angle=270]{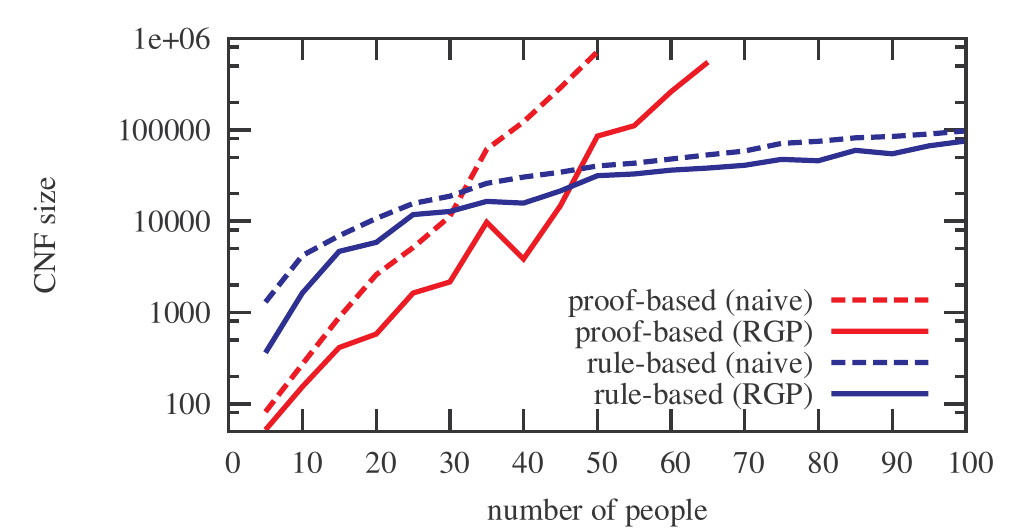}}
\subfigure[runtime of exact inference (compilation+traversal)]{\includegraphics[height=8.0cm,width=3.4cm,angle=270]{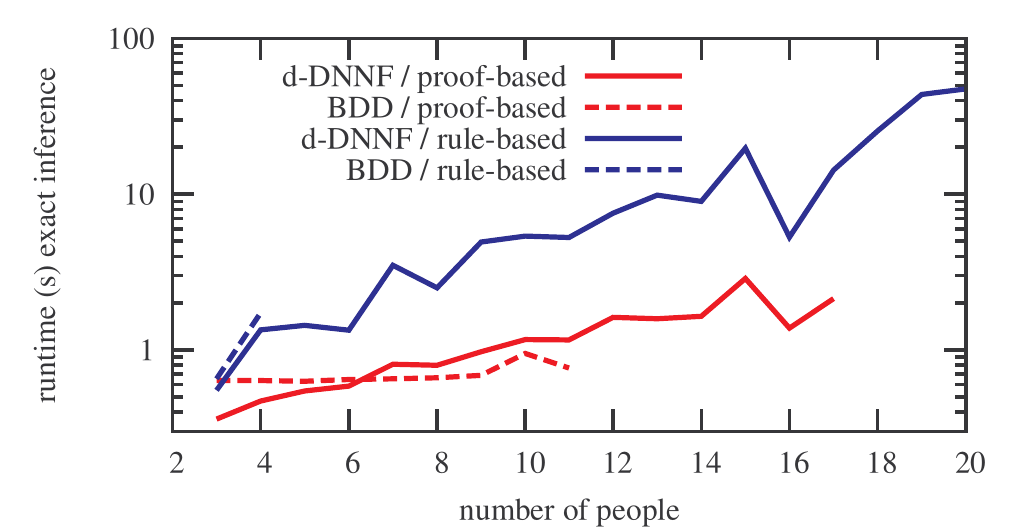}} 
\subfigure[normalized negative log-likelihood of MC-SAT (lower is better)]{\includegraphics[height=8.0cm,width=3.4cm,angle=270]{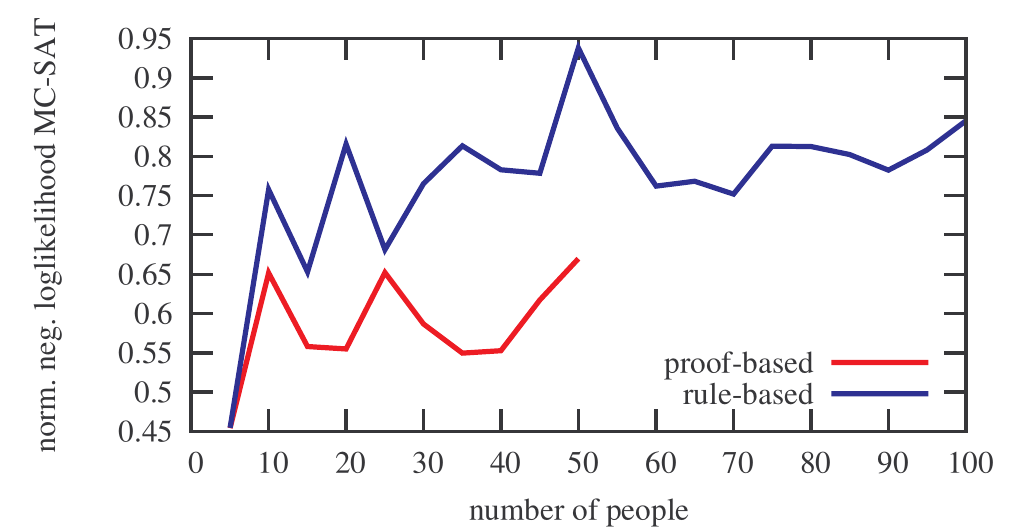}} 
\subfigure[number of samples drawn by MC-SAT]{\includegraphics[height=8.0cm,width=3.4cm,angle=270]{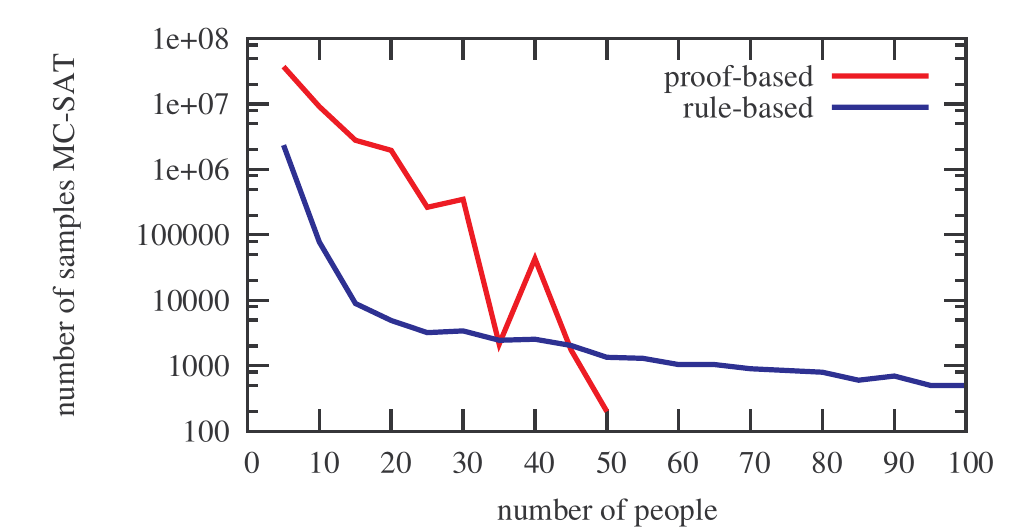}} 
\caption{Results for \emph{Smokers} as a function of domain size. (When the curve for an algorithm ends at a particular domain size, this means that the algorithm is intractable beyond that size.)} \label{fig:smokers}
\end{figure*}

\begin{figure*}
\subfigure[runtime of conversion to a weighted Boolean formula]{\includegraphics[height=8.0cm,width=3.4cm,angle=270]{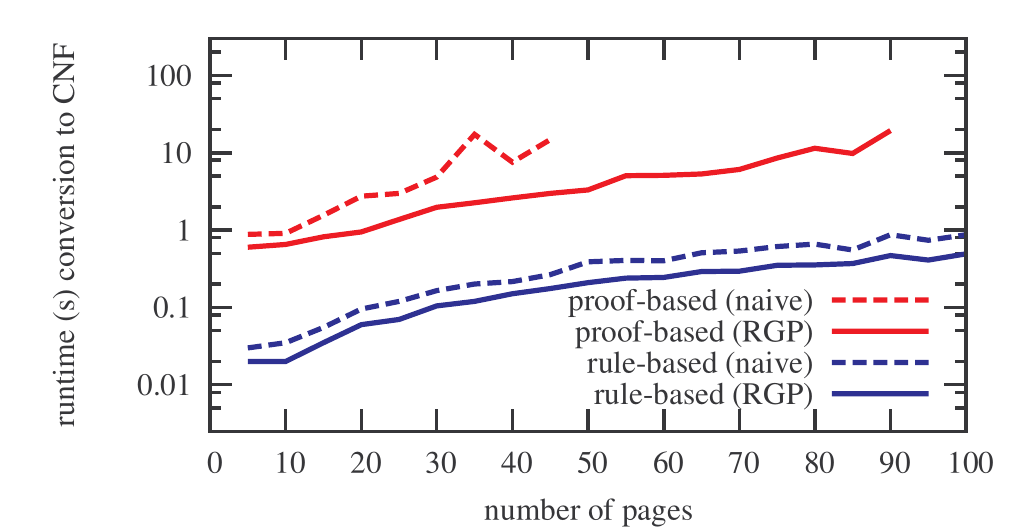}} 
\subfigure[size of the Boolean formula (number of clauses in the CNF)]{\includegraphics[height=8.0cm,width=3.4cm,angle=270]{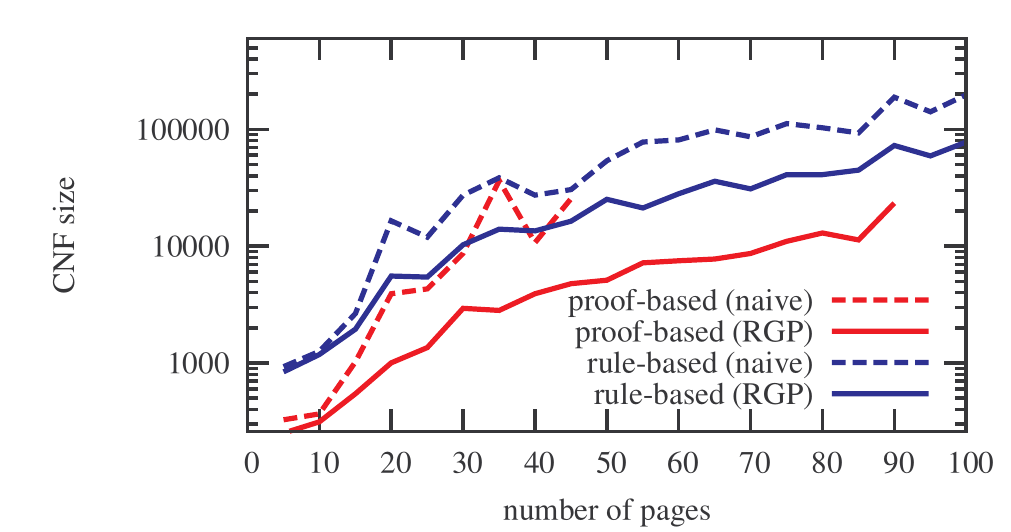}}
\subfigure[runtime of exact inference (compilation+traversal)]{\includegraphics[height=8.0cm,width=3.4cm,angle=270]{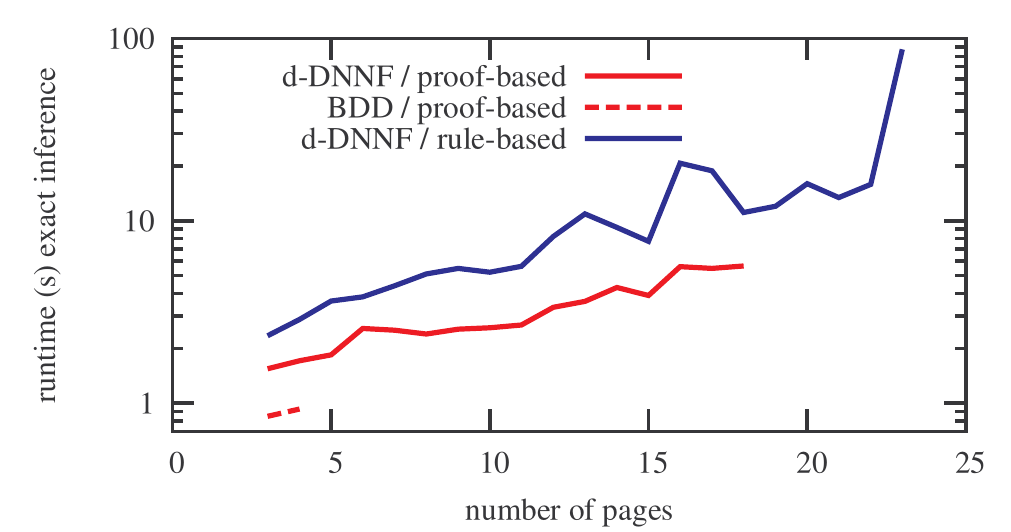}} 
\subfigure[normalized negative log-likelihood of MC-SAT (lower is better)]{\includegraphics[height=8.0cm,width=3.4cm,angle=270]{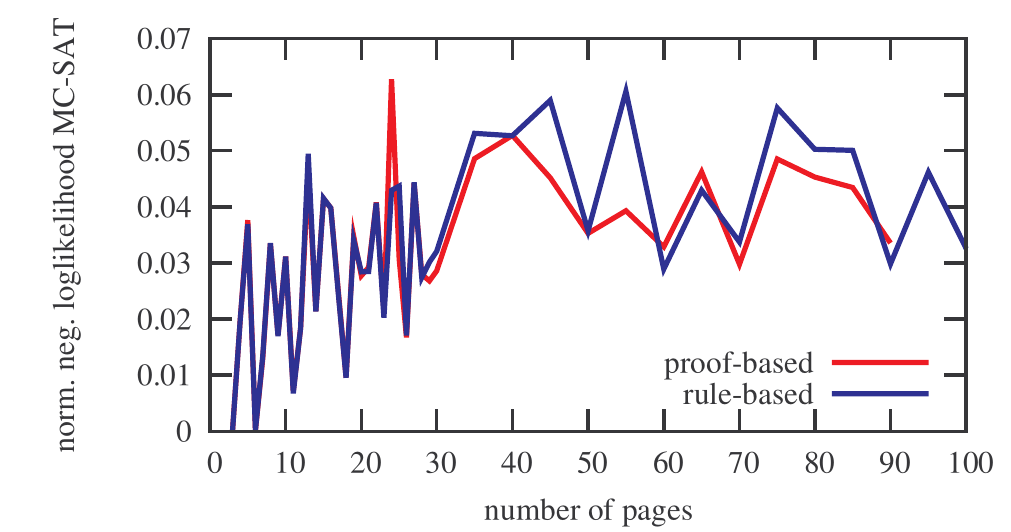}} 
\subfigure[number of samples drawn by MC-SAT]{\includegraphics[height=8.0cm,width=3.4cm,angle=270]{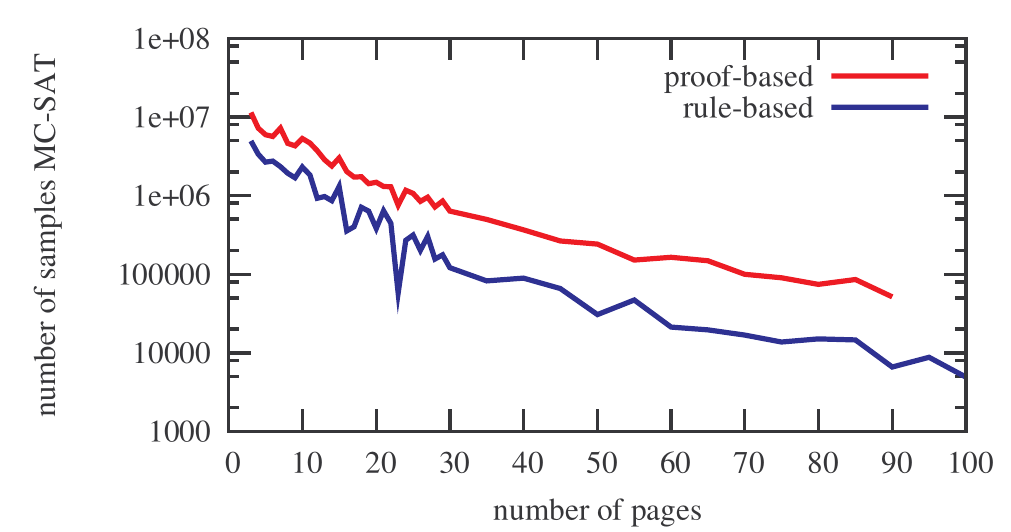}} 
\caption{Results for \emph{WebKB} as a function of domain size. (When the curve for an algorithm ends at a particular domain size, this means that the algorithm is intractable beyond that size.)} \label{fig:webkb}
\end{figure*}

\subsubsection{Q1 - Influence of the Grounding Algorithm}

Question {\bf Q1} is: is working with the relevant rather than the complete ground program more efficient? 
To answer this question, we measured the time needed for grounding, the size of the resulting ground program, and the size of the Boolean formula derived from this program.

\emph{The grounding step.} The idea behind the relevant ground program (RGP) is to reduce the grounding by pruning clauses that are irrelevant or inactive w.r.t. the queries and evidence. Our setup 
is such that all clauses are relevant. Hence, the only reduction comes from pruning inactive clauses (that have a false evidence literal in the body). The effect of this pruning is small: on average the size of the ground program is reduced by 17\% (results not shown).

\emph{Implications on the conversion to a Boolean formula.} 
The proof-based conversion becomes intractable (i.e., takes prohibitively long) for large domain sizes, but the size where this happens is significantly larger when working on the RGP instead of on the complete grounding (see Fig.~\ref{fig:smokers}a/\ref{fig:webkb}a). Also the size of the Boolean formula is reduced significantly by using the RGP (up to a 90\% reduction in number of clauses in the CNF, Fig.~\ref{fig:smokers}b/\ref{fig:webkb}b). The reason why a 17\% reduction of the program can yield a 90\% reduction of the corresponding formula is that loops in the program cause a `blow-up' of the formula. Removing only a few rules in the ground program can already break loops and make the formula significantly smaller. Note that the proof-based conversion suffers from this blow-up more than the rule-based conversion does.

Computing the grounding is always very fast, both for the RGP and the complete grounding (milliseconds on Smokers; around 1s for WebKB). Hence, as an answer to question {\bf Q1}, we conclude that using the RGP instead of the complete grounding is beneficial and comes at almost no computational cost. Hence, from now on we always use the RGP.

\subsubsection{Q2 - Influence of the Conversion Algorithm}
\label{sec:MCSATexp}

Question \textbf{Q2} is: which of the two considered algorithms for converting the ground program to a Boolean formula performs best? Recall that we have seen a rule-based and a proof-based conversion (Section~\ref{sec:lp2bool}). To answer this question, we measure the time of the conversion process, the size of the resulting formula, and how efficient this formula is for inference.

\emph{The conversion step.} The proof-based algorithm, by its nature, does more effort to convert the program into a compact formula. This has implication on the scalability of the algorithm: on small domains the algorithm is fast, but on larger domains it becomes intractable (Fig.~\ref{fig:smokers}a/\ref{fig:webkb}a). In contrast, the rule-based algorithm is able to deal with all considered domain sizes and is always fast (runtime at most 0.5s). A similar trend holds in terms of the size of the formula. For small domains, the proof-based algorithm generates smaller formulas than the rule-based algorithm, but for larger domains the opposite holds (Fig.~\ref{fig:smokers}b/\ref{fig:webkb}b).

\emph{Implications on inference.} The ultimate question is how efficient the formulas of the different conversions are for subsequent inference. We discuss this for exact inference in the next section; here we focus on approximate inference. We use the \mbox{MC-SAT} inference algorithm (Section~\ref{sec:MARGsolution}) as a tool to evaluate how efficient the different formulas are for inference by running MC-SAT on the two types of formulas and measuring 
the quality of the estimated marginals. Evaluating the quality of approximate marginals is non-trivial when computing true marginals is intractable. We use the same solution as the original MC-SAT paper: we let MC-SAT run for a fixed time (10 minutes) and measure the quality of the estimated marginals as the likelihood of the `query ground truth' according to these estimates; see \citeN{MCSAT}.

On domain sizes where the proof-based algorithm is still tractable, inference results are better with the proof-based formula than with the rule-based formula (see Fig.~\ref{fig:smokers}d, and to a smaller extent Fig.~\ref{fig:webkb}d). This is because the proof-based formulas are more compact and hence more samples can be drawn in the given time (Fig.~\ref{fig:smokers}e/\ref{fig:webkb}e).

As an answer to question {\bf Q2}, we conclude  that for smaller domains the proof-based algorithm is preferable because of the smaller formulas. On larger domains, the rule-based algorithm should be used.

\subsubsection{Q3 - Influence of the Inference Algorithm}
\label{sec:exp_BDD_dDNNF}

For exact inference, our approach consists of knowledge compilation, with either d-DNNFs or BDDs. Question {\bf Q3} is: which of the two considered approaches, d-DNNFs or BDDs, performs best? To answer this question, we increase the domain size up to the point where inference (doing the compilation to d-DNNF or BDD) becomes intractable. 
It is useful to distinguish between compilation of rule-based and proof-based formulas.\footnote{In the PLP literature, BDDs have almost exclusively been used for proof-based formulas \cite{ProbLogIJCAI07,GutmannECML11}. Compiling our proof-based formulas to BDDs yields exactly the same BDDs as used by \citeN{GutmannECML11}. In the special case of a single query and no evidence, this also equals the BDDs used \citeN{ProbLogIJCAI07}.}  

\emph{Proof-based formulas.} On the Smokers domain, BDDs perform relatively well, but they are nevertheless clearly outperformed by the d-DNNFs (Fig.~\ref{fig:smokers}c). On WebKB, the difference is even larger: BDDs are only tractable on domains of size 3 or 4, while d-DNNFs reach up to size 18 (Fig.~\ref{fig:webkb}c). When BDDs become intractable, this is mostly due to memory problems.\footnote{It might be surprising that BDDs, which are the state-of-the-art in PLP, do not perform better. However, one should keep in mind that we are using BDDs for \emph{exact} inference here. BDDs are also used for approximate inference, one simply compiles an \emph{approximate formula} into a BDD \cite{ProbLogIJCAI07}. The same can be done with d-DNNFs, and we again expect improvement over BDDs.}


\emph{Rule-based formulas.} As seen before, these formulas are less compact than the proof-based formulas (at least for those domain sizes where exact inference is feasible). The results clearly show that the d-DNNFs are much better at dealing with these non-compact formulas than the BDDs are. Concretely, the d-DNNFs are still tractable up to reasonable sizes. In contrast, using BDDs on these rule-based formulas is nearly impossible: on Smokers the BDDs only solve size 3 and 4, on WebKB they even do not solve any of the inference tasks on rule-based formulas.     

As an answer to question {\bf Q3}, we conclude that the use of d-DNNFs pushes the limit of exact MARG inference significantly further as compared to BDDs, which were the standard in PLP. 

\subsubsection{Q4 - Computing Success Probabilities with ProbLog2}

Question {\bf Q4} is: when using the `classical' ProbLog setting of computing success probabilities, does ProbLog2 (our new ProbLog implementation) outperform ProbLog1 (the previous ProbLog implementation)? 

For ProbLog1 we use the default parameter settings and we table the $\mathit{path/2}$ predicate. For ProbLog2 we use the proof-based conversion and we compile to d-DNNF. These settings are motivated by our previous conclusions, which show that the proof-based conversion works well on programs that are small enough to allow for exact inference (as we do here) and that d-DNNFs are superior to BDDs. 

\begin{figure*}
\includegraphics[width=11cm,height=6cm]{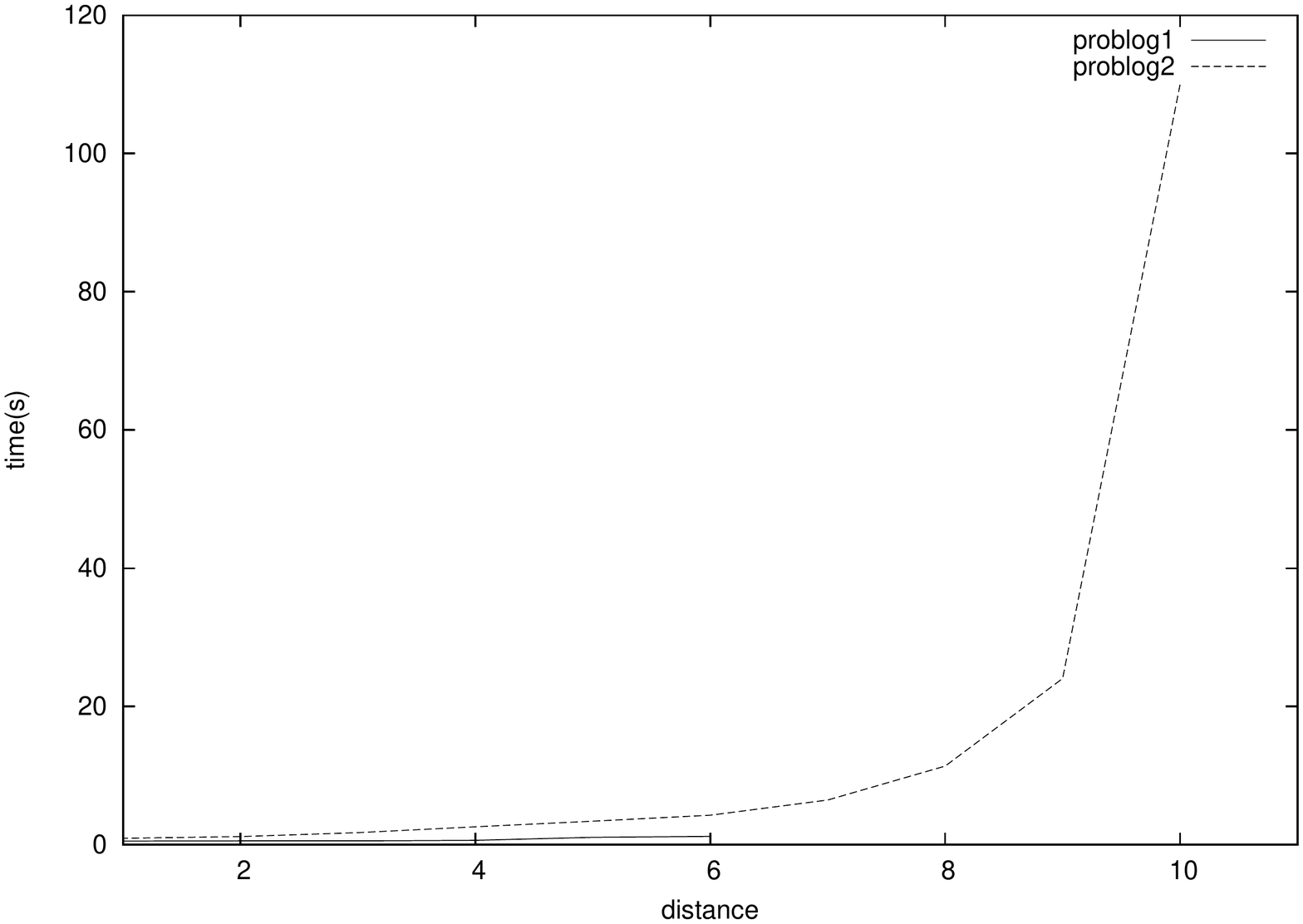}
\caption{Runtime of ProbLog1 and ProbLog2 on the probabilistic grid query, as a function of the distance $16-i$ between the start and end node. (When the curve for an algorithm ends at a particular point, this means that the algorithm is intractable beyond that point.)} \label{fig:alzheimermedium}
\end{figure*}

As explained before (Section~\ref{sec:inf_setup}), we ask the query \verb;path(n_i_i,n_16_16);, where we vary $i$ from 15 to 1. The smaller $i$, the larger the `distance' $16-i$ between the start and end node, and hence the harder the problem. Figure~\ref{fig:alzheimermedium} shows the measured runtimes for ProbLog1 and ProbLog2.

The results show that ProbLog2 scales better than ProbLog1. ProbLog1 is tractable up to distance 6. From distance 7 onwards, it becomes intractable, i.e., it incurs a time-out. We have put the time limit on 300 seconds and repeated every experiment 10 times. For distance 7, all 10 repetitions timed-out, while for distance 6 the average runtime was only 1.2 seconds.\footnote{To verify that the measurement for distance 7 is not a glitch, we also tried distance 8 and further, but ProbLog1 consistently timed-out for all of these.} This shows that the runtime of ProbLog1 explodes beyond distance 6. In contrast, ProbLog2 is tractable up to distance 10. For distance 10, all 10 repetitions finish in time, taking on average 110 seconds. For distance 11, only 5 out of 10 repetitions still finish in time. From distance 12 onwards, all 10 repetitions time out.  





\subsubsection{Q5 - Ability to Learn the Original Probabilities}

Question {\bf Q5} is: when learning from data generated from a known program, can we recover the parameters of the original program given a reasonable amount of data? We answer this question by generating data from the given Smokers program (Section~\ref{sec:exp:data}), applying our learning algorithm to this data, and measuring the difference between the learned probabilities and those in the original program. We measure this difference in two ways. First, we use the mean absolute error (MAE) between both sets of probabilities. Second, we use the Kullback-Leibler(K-L)-divergence, a measure of similarity between a `true' probability distribution (the one of the original program) and an `approximating' distribution (the one of the learned program). ProbLog allows for an efficient calculation of the K-L-divergence because of the independence of the probabilistic facts; see~\ref{apd:kldiv}.

\begin{figure*}
\subfigure[4 persons]{\includegraphics[height=4.5cm,width=8.6cm]{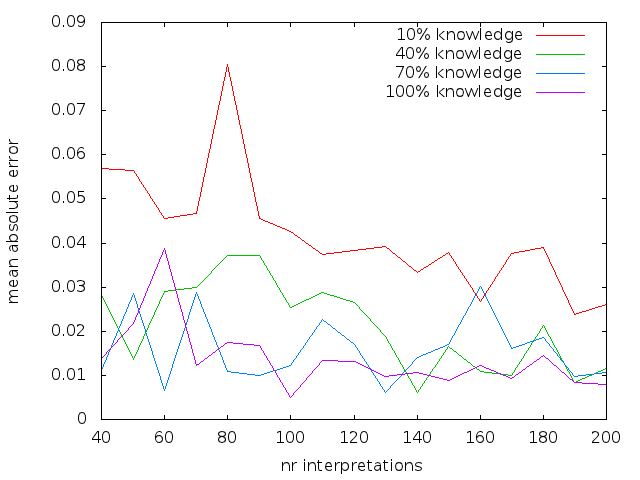}} 
\subfigure[5 persons]{\includegraphics[height=4.5cm,width=8.6cm]{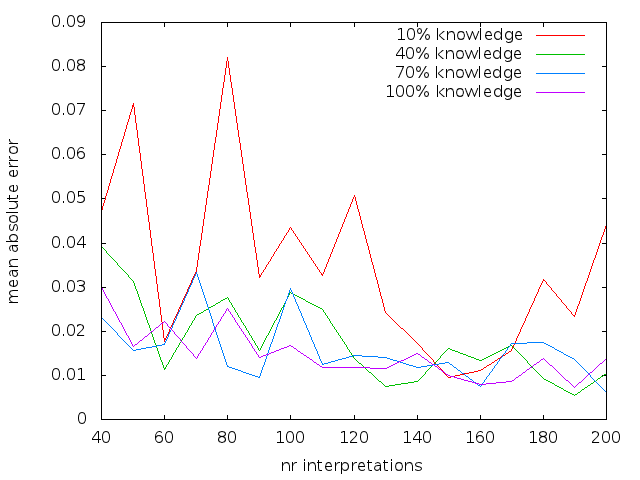}}
\subfigure[6 persons]{\includegraphics[height=4.5cm,width=8.6cm]{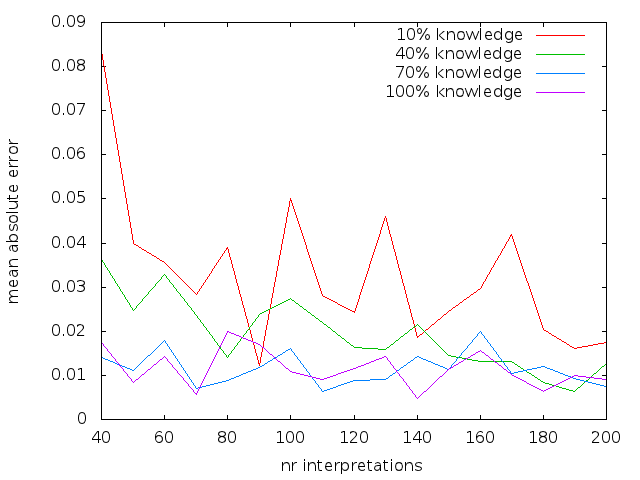}} 
\subfigure[7 persons]{\includegraphics[height=4.5cm,width=8.6cm]{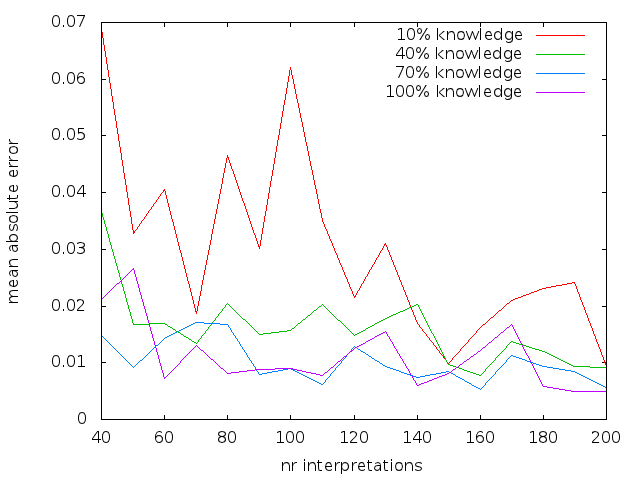}} 
\caption{Mean absolute error (MAE, lower is better) when learning from Smokers data with $10\%$, $40\%$, $70\%$ and $100\%$ knowledge of the possible world.} \label{fig:mae}
\end{figure*}

\begin{figure*}
\subfigure[4 persons]{\includegraphics[height=4.5cm,width=8.6cm]{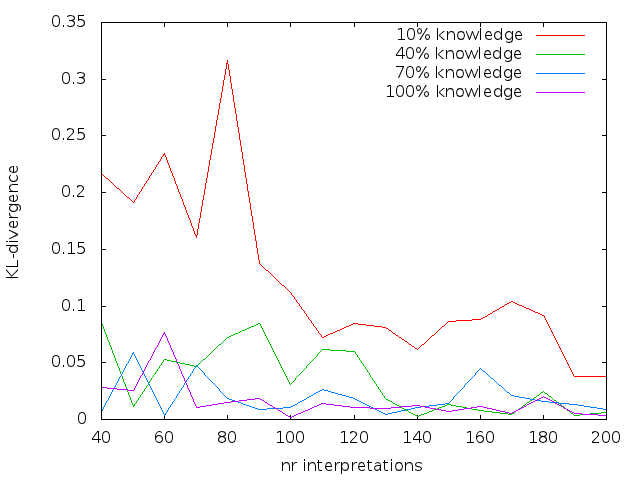}} 
\subfigure[5 persons]{\includegraphics[height=4.5cm,width=8.6cm]{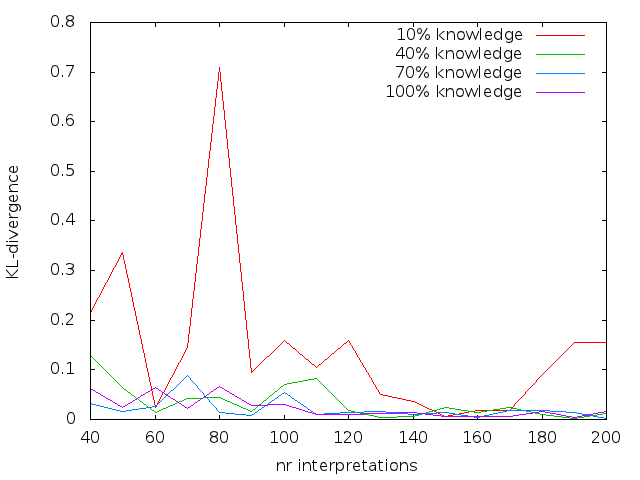}} 
\subfigure[6 persons]{\includegraphics[height=4.5cm,width=8.6cm]{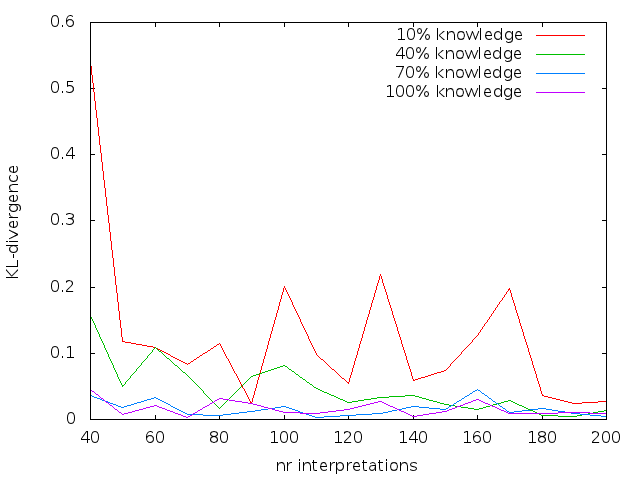}} 
\subfigure[7 persons]{\includegraphics[height=4.5cm,width=8.6cm]{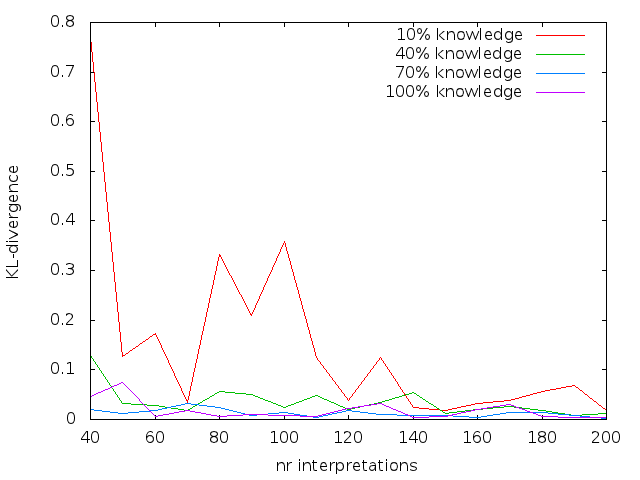}} 
\caption{K-L-divergence (lower is better) when learning from Smokers data with $10\%$, $40\%$, $70\%$ and $100\%$ knowldege of the possible world.} \label{fig:kldiv}
\end{figure*}

Both the MAE (Figure~\ref{fig:mae}) and the K-L-divergence (Figure~\ref{fig:kldiv}) show that LFI-ProbLog can learn the original probabilities: both MAE and K-L-divergence approach zero when more examples are given. The $100\%$ knowledge line shows the optimal way of calculating the probabilities, given the interpretations. The remaining cases, $10\%$, $40\%$ and $70\%$ show that the quality of the approximations, as expected, drops when more atoms become unobserved. However, the approximations remain of good quality. Hence we can conclude that LFI-ProbLog is capable of recovering the original probabilities and is robust against missing values. When we compare the figures for the different domain sizes, we see that the results are independent of the number of persons in the domain.

\subsubsection{Q6 - Learning of real-world data}

Question \textbf{Q6} is: when learning from real-world data, can we obtain results comparable to the ones obtained with a state-of-the-art system? To answer this question, we compare LFI-ProbLog with the Alchemy system for Markov Logic, running four-fold cross validation on the WebKB dataset. Table \ref{webkb_results} shows the negative log-likelihood obtained with LFI-ProbLog and Alchemy on the test-sets of the four folds. In the case of Alchemy, we report two results: `Alchemy' stands for using the system with its default parameters, `Alchemy*' stands for Alchemy with a modified setting that puts a very strong prior on the weights (prior around zero, with standard deviation 0.1 instead of the default 100).\footnote{This modified setting was recommended to us by the Alchemy developers (personal communication with Daniel Lowd).}
  
\begin{table}
  \label{webkb_results}
  \centering
    \begin{tabular}{l|ccc}
	Test Set & LFI-Problog & Alchemy* & Alchemy \\
      \hline
      Cornell & 1309.72 & 613.37 & 1603.31\\
      Texas & 1210.51 & 640.56 & 1075.75 \\
      Washington & 646.39 & 622.55 & 1420.87\\
      Wisconsin & 1033.90 & 783.51 & 3479.04\\
      \hline
  \end{tabular}
  \caption{Negative Log-Likelihood (lower is better) on the WebKB learning experiment.}
\end{table}

LFI-ProbLog outperforms Alchemy with its default settings on three of the four folds. However, Alchemy with the strong prior (Alchemy*) outperforms LFI-ProbLog on all four folds. We conclude that LFI-ProbLog is competitive with Alchemy, but parameter tuning can have a large impact. These results illustrate the importance of setting a suitable prior when learning. This is a topic that we have not yet explored in detail for LFI-ProbLog but that we plan to study in future research.


\section{Conclusion}

The contributions of this paper are threefold.

First, we have introduced a two-step procedure for MPE and MARG inference in general probabilistic logic programs. In a first step it generates a weighted Boolean formula that captures all relevant information about a specific query, evidence and probabilistic logic program. This step relies on well-known conversion techniques from logic 
programming. The second step then invokes well-known solvers (for instance for WMC and weighted  MAX-SAT)
on the generated weighted formula.  

The resulting inference procedure is akin to that employed by \citeN{DarwicheBook} and others \cite{Park02,Sang05} for probabilistic graphical models (where   
many inference problems are also cast in terms of weighted Boolean formulas) but adapted to the much more expressive class of probabilistic logic programs. 
Our conversion-based approach is advantageous because it allows us to employ a wide range of well-known and optimized solvers on the weighted formula, essentially giving us ``inference algorithms for free''.
Furthermore, the approach also improves upon the state-of-the-art in probabilistic logic programming, where one  has typically focussed on inference with a single query atom and no evidence 
(cf.\ Section~\ref{sec:tasks}), often by using BDDs. By using d-DNNFs instead of BDDs, we obtained speed-ups that push the limit of exact MARG inference significantly further.   

Second, we have developed an Expectation-Maximization approach to learning probabilistic logic programs from interpretations. This approach employs our novel inference procedures in the expectation step. The learning from interpretation setting is akin to that used in the graphical model and Statistical Relational Learning (SRL) communities.  

Third, the two approaches have been incorporated in a novel implementation of the PLP language ProbLog,
which unlike its previous implementation in YAP-Prolog \cite{KimmigTPLP11}, is closer to that of answer set programming systems than to Prolog systems,

Overall our approach provides new insights into the relationships between PLP, graphical models and SRL. As one immediate outcome, we pointed out a conversion of  probabilistic logic programs to ground Markov Logic, which allowed us to apply MC-SAT to PLP inference. This contributes to further bridging the gap between PLP and the field of SRL.


\appendix

\section{Proofs}

In this appendix we give the proofs of Theorem~1 to~3.

\subsection*{Proof of Theorem 1}

To prove Theorem 1 we first give necessary some lemma's. We use $\mathit{pruneInactive}(L,$ $\mathbf{E}=\mathbf{e})$ to denote the result of removing from a ground program $L$ all rules that are inactive under the evidence $\mathbf{E}=\mathbf{e}$.

\begin{lemma} 
Let $L$ be a ground normal logic program and let $L' = \mathit{pruneInactive}(L,\mathbf{E}=\mathbf{e})$. 
For each world/interpretation $\omega$ that is consistent with the evidence $\mathbf{E}=\mathbf{e}$ it holds:
\begin{itemize}
\item [a)] for each subset $A$ of atoms: $A$ is an unfounded set with respect to $\omega$ under program $L$ if and only if it is so under program $L'$, and
\item [b)] $\omega$ is the well-founded model of $L$ if and only if it is the well-founded model of $L'$. 
\end{itemize}
\end{lemma}

\noindent \emph{Proof:} 

\emph{Part a:} We use the notion of \emph{unfounded set} see Definition 3.1 in \citeN{wellfounded91}. We prove both directions of the `if and only if'.
\begin{itemize}
\item If $A$ is an unfounded set with respect to $\omega$ under program $L$, then this also holds under program $L'$:\\
 The definition of unfounded set imposes a certain condition on each rule in the program whose head is in the set $A$, we refer to this as the unfounded rule condition. If we know that this condition holds for all such rules in $L$, then it also holds for all such rules in $L'$, because the latter set of rules is a subset of the former ($L'$ is the result of removing inactive rules from $L$).
\item If $A$ is an unfounded set with respect to $\omega$ under program $L'$, then this also holds under program $L$: \\
The `if' part of this `if-then' implies that the unfounded rule condition holds for all rules in $L'$, so to prove the `then' part we only need to show that the unfounded rule condition also holds for all rules in $L \setminus L'$ (i.e., for all rules in $L$ that were removed because of being inactive under the evidence). Every rule $r \in L \setminus L'$ contains at least one atom in its body that is false according to the evidence (that is what made $r$ inactive). Since this lemma applies only to worlds $\omega$ that are consistent with the evidence, we have for every such world $\omega$: every rule $r \in L \setminus L'$ contains in its body at least one atom false in $\omega$. This is a sufficient condition to make the rule $r$ an unfounded rule, see condition `1.' in Definition 3.1 in \citeN{wellfounded91}.
\end{itemize}

\emph{Part b:} We can now use \emph{Part a} to prove that, for every evidence-consistent world $\omega$, $\omega$ is the well-founded model (WFM) of $L$ if and only if it is the WFM of $L'$. 

The WFM is the fixed point of the $W_P$ operator \cite{wellfounded91}. For a program $L$, this operator is defined as $W_L(\omega) = T_L(\omega) \cup \lnot U_L(\omega)$, see Definition 3.3 in \citeN{wellfounded91}. We prove below that, for every evidence-consistent world $\omega$: $T_L(\omega)=T_{L'}(\omega)$ and $U_L(\omega)=U_{L'}(\omega)$. Hence, $W_L(\omega)=W_{L'}(\omega)$, hence their fixed points are identical, hence \emph{Part b} holds.
\begin{itemize}
\item For every evidence-consistent world $\omega$: $T_L(\omega)=T_{L'}(\omega)$ \\
$L$ consists of all rules in $L'$ plus some rules that are inactive under the evidence. For each evidence-consistent $\omega$, the bodies of the inactive rules in $L$ are false under $\omega$ and hence these rules cannot `fire'. Hence these rules play no role in the execution of the $T_P$ operator on $\omega$. Hence $T_L(\omega)=T_{L'}(\omega)$.

\item For every evidence-consistent world $\omega$: $U_L(\omega)=U_{L'}(\omega)$: \\ 
$U_L(\omega)$ is the greatest unfounded set with respect to (wrt) $\omega$, which is defined as the union of all unfounded sets wrt $\omega$, see Definition 3.2 in \citeN{wellfounded91}. \emph{Part a} says that any subset $A$ of atoms is an unfounded set wrt $\omega$ under program $L$ if and only if it is so under program $L'$. Hence $U_L(\omega)=U_{L'}(\omega)$.
\end{itemize}
$\Box$

%

\begin{lemma} 
Let $L$ be a ground ProbLog program and let $L' = \mathit{pruneInactive}(L,\mathbf{E}=\mathbf{e})$. 
Then $MOD_{\mathbf{E}=\mathbf{e}}(L) = MOD_{\mathbf{E}=\mathbf{e}}(L')$.
\end{lemma}

\noindent \emph{Proof:} $MOD_{\mathbf{E}=\mathbf{e}}(L)$ is defined as the set of all worlds $\omega$ that are consistent with the evidence $\mathbf{E}=\mathbf{e}$ and are models of the ProbLog program $L$, i.e., for which there exists a total choice $C$ and $WFM(C \cup R) = \omega$, with $R$ the rules in $L$ and $WFM()$ the well-founded model. The previous lemma implies that, for every $\omega$ consistent with the evidence, removing inactive rules from a given logic program does not alter whether or not $\omega$ is the WFM of that program or not. In other words: $\omega \in MOD_{\mathbf{E}=\mathbf{e}}(L)$ if and only if $\omega \in MOD_{\mathbf{E}=\mathbf{e}}(L')$. Hence $MOD_{\mathbf{E}=\mathbf{e}}(L) = MOD_{\mathbf{E}=\mathbf{e}}(L')$. $\Box$

\begin{lemma} 
Let $L$ be a ground ProbLog program and let $L' = \mathit{pruneInactive}(L,\mathbf{E}=\mathbf{e})$. 
Then $P_L(Q \mid \mathbf{E}=\mathbf{e}) = P_{L'}(Q \mid \mathbf{E}=\mathbf{e})$
\end{lemma}

\noindent \emph{Proof:}
We prove the stronger condition $\forall \omega: P_L(\omega \mid \mathbf{E}=\mathbf{e}) = P_{L'}(\omega \mid \mathbf{E}=\mathbf{e})$
The conditional probability $P_L(\omega \mid \mathbf{E}=\mathbf{e})$ of an interpretation $\omega$ according to program $L$ is:
\begin{itemize}
\item (\underline{case1}) if $\omega \in MOD_{\mathbf{E}=\mathbf{e}}(L)$ then
\[P_L(\omega \mid \mathbf{E}=\mathbf{e}) = \frac{P_L(\omega, \mathbf{E}=\mathbf{e})}{P_L(\mathbf{E}=\mathbf{e})}.\]
Since $\omega$ agrees with $\mathbf{E}=\mathbf{e}$, we have that the combined assignment $\omega, \mathbf{E}=\mathbf{e}$ is simply equal to $\omega$. Hence:
\begin{equation}\label{eq:condprobdef}
P_L(\omega \mid \mathbf{E}=\mathbf{e}) = \frac{P_L(\omega)}{P_L(\mathbf{E}=\mathbf{e})} = \frac{P_L(\omega)}{\sum_{\omega' \in MOD_{\mathbf{E}=\mathbf{e}}(L)}P_L(\omega')}. 
\end{equation}
\item (\underline{case2}) if $\omega \notin MOD_{\mathbf{E}=\mathbf{e}}(L)$ then $P_L(\omega \mid \mathbf{E}=\mathbf{e}) = 0$.
\end{itemize}
We now prove that for every $\omega$, $P_L(\omega \mid \mathbf{E}=\mathbf{e})=P_{L'}(\omega \mid \mathbf{E}=\mathbf{e})$. The proof consists of two parts.
\begin{enumerate}
\item We need to prove that we are in case1 under $L$ if and only if we are in case1 under $L'$. In other words: for every $\omega$: $\omega \in MOD_{\mathbf{E}=\mathbf{e}}(L)$ if and only if $\omega \in MOD_{\mathbf{E}=\mathbf{e}}(L')$. This follows from the previous lemma.
\item We need to prove that if we are in case1 (i.e.\ if $\omega \in MOD_{\mathbf{E}=\mathbf{e}}(L)$), then the conditional probability given by the fraction in Equation~\ref{eq:condprobdef} is the same under $L$ as under $L'$. 
  \begin{itemize}
  \item The \underline{numerator} is the same under $L$ and $L'$. This can be seen as follows. For any $\omega \in MOD_{\mathbf{E}=\mathbf{e}}(L)$, the probability $P(\omega)$ is by definition equal to the probability of $\omega$'s total choice. The ProbLog programs $L$ and $L'$ differ in their rules, but they have exactly the same probabilistic facts and hence determine the same probability distribution over total choices. Hence $P(\omega)$ is the same under $L$ as under $L'$.
  \item The sum in the \underline{denominator} is also the same under $L$ and $L'$. This can be seen as follows. First, the set $MOD_{\mathbf{E}=\mathbf{e}}(L)$ over which the sum ranges is the same under $L$ as under $L'$ because of the above lemma. Second, each term in the sum is the same under L as under L', i.e.\  for every $\omega \in MOD_{\mathbf{E}=\mathbf{e}}(L)$ the probability $P(\omega)$ is the same under $L$ as under $L'$ (because of the same reasoning as for the numerator).
  \end{itemize}
\end{enumerate}
This concludes the proof. $\Box$

\setcounter{theorem}{0}
\begin{theorem} 
Let $L$ be a ProbLog program and let $L_g$ be the relevant ground program for $L$ with respect to $\mathbf{Q}$ and $\mathbf{E}=\mathbf{e}$. Then $P_L(Q \mid \mathbf{E}=\mathbf{e}) = P_{L_g}(Q \mid \mathbf{E}=\mathbf{e})$.
\end{theorem}

\noindent \emph{Proof:} It follows from the grounding semantics of ProbLog that replacing the original program $L$ by its full grounding (w.r.t.\ the Herbrand base) $L_{full}$ preserves the distribution, i.e., $P_L(\mathbf{Q} \mid \mathbf{E}=\mathbf{e}) = P_{L_{full}}(\mathbf{Q} \mid \mathbf{E}=\mathbf{e})$. The relevant ground program $L_g$ differs from $L_{full}$ only in that it does not contain inactive rules (with respect to $\mathbf{E}=\mathbf{e}$) or irrelevant rules (with respect to $\mathbf{Q} \cup \mathbf{E}$). The lemma above states that removing inactive rules preserves the distribution $P(\mathbf{Q} \mid \mathbf{E}=\mathbf{e})$. Removing irrelevant rules also preserves this distribution; this can be seen as follows. The probability of an atom being true can be determined from all proofs of the atom and the probabilities of the probabilistic facts appearing in these proofs, see \citeN{ProbLogIJCAI07}. Irrelevant rules are - by definition - rules that are not used in any proof of any atom in $\mathbf{Q} \cup \mathbf{E}$. Hence omitting such irrelevant rules does not alter the distribution $P(\mathbf{Q}, \mathbf{E})$. Hence, also the distribution $P(\mathbf{Q} \mid \mathbf{E}=\mathbf{e})$ is preserved because $P(\mathbf{Q} \mid \mathbf{E}=\mathbf{e})$ can be defined in terms of $P(\mathbf{Q}, \mathbf{E})$, i.e., $P(\mathbf{Q} \mid \mathbf{E}=\mathbf{e}) = \frac{P(\mathbf{Q}, \mathbf{E}=\mathbf{e})}{P(\mathbf{E}=\mathbf{e})} = \frac{P(\mathbf{Q}, \mathbf{E}=\mathbf{e})}{\sum_{\mathbf{q}} P(\mathbf{Q}=\mathbf{q}, \mathbf{E}=\mathbf{e})}$.


\subsection*{Proof of Theorem 2}

\begin{theorem}
Let $L_g$ be the relevant ground program for some ProbLog program with respect to $\mathbf{Q}$ and \mbox{$\mathbf{E}=\mathbf{e}$}. Let $MOD_{\mathbf{E=e}}(L_g)$ be those models in $MOD(L_g)$ that are consistent with the evidence $\mathbf{E=e}$. Let $\varphi$ denote the formula and $w(.)$ the weight function of the weighted formula derived from $L_g$. Then:
\begin{itemize}
\item[-] \textbf{(model equivalence)} $SAT(\varphi) = MOD_{\mathbf{E=e}}(L_g)$, 
\item[-] \textbf{(weight equivalence)} $\forall \omega \in SAT(\varphi)$: $w(\omega) = P_{L_g}(\omega)$, i.e., the weight of $\omega$ according to $w(.)$ is equal to the probability of $\omega$ according to $L_g$.
\end{itemize}
\end{theorem}

\noindent \emph{Proof:} The proof consists of two parts. 

\noindent \textbf{Model equivalence.} Consider Lemma 1 (Section 5.2). The lemma is about the formula $\varphi_r$ that captures the rules but not yet the evidence. The lemma states that $SAT(\varphi_r)=MOD(L_g)$. The present theorem is about the formula $\varphi = \varphi_r \land \varphi_e$, where $\varphi_e$ captures the evidence. The effect of adding $\varphi_e$ to the formula is that all worlds not consistent with the evidence are ruled out. Hence $SAT(\varphi_r \land \varphi_e) = MOD_{\mathbf{E}=\mathbf{e}}(L_g)$.

\noindent \textbf{Weight equivalence.} 
Weight equivalence says that the probability of every model (according to $L_g$) is equal to the weight of the model (according to our weight function $w(.)$). This follows from the way the probability and the weight function are defined.
\begin{itemize}
\item The \textbf{probability} of a model of a ProbLog program, according to the distribution semantics, is the probability of the underlying total choice, which in turn is defined as the product of probabilities of each of the atomic choices. Formally, the probability of a model $\omega$ is: 
\[P(\omega) = \prod_{a \in \mathit{PA}^{+}(\omega)} p(a) \prod_{a \in \mathit{PA}^{-}(\omega)} p(\lnot a) = \prod_{a \in \mathit{PA}^{+}(\omega)} p(a) \prod_{a \in \mathit{PA}^{-}(\omega)} (1-p(a)),\] 
with $\mathit{PA}^{+}(\omega)$ (respectively $\mathit{PA}^{-}(\omega)$) being the set of all ground probabilistic atoms that are true (resp.\ false) in $\omega$ and $p(.)$ denoting the probability distribution specified by the probabilistic facts.
\item The \textbf{weight} of a world $\omega$ according to our weight function is the product of the weights of all literals $l$ constituting the world/interpretation $\omega$:
\[w(\omega) = \prod_{l \in \omega} w(l) .\]
The literals/atoms in $\omega$ fall into four groups: 
probabilistic atoms that are true in $\omega$ (denoted $\mathit{PA}^{+}(\omega)$, non-probabilistic or derived atoms that are true in $\omega$ (denoted $\mathit{DA}^{+}(\omega)$), and similar for the atoms that are false in $\omega$ ($\mathit{PA}^{-}(\omega)$ and $\mathit{DA}^{-}(\omega)$). Hence:
\[w(\omega) = \prod_{l \in \omega} w(l) = \prod_{a \in \mathit{PA}^{+}(\omega)} w(a) \prod_{a \in \mathit{PA}^{-}(\omega)} w(\lnot a) \prod_{a \in \mathit{DA}^{+}(\omega)} w(a) \prod_{a \in \mathit{DA}^{-}(\omega)} w(\lnot a) .\]
By definition of the weight function, the weight of an atom $a \in \mathit{PA}^{+}(\omega)$ is $p(a)$, the weight of $a \in \mathit{PA}^{-}(\omega)$ is $1-p(a)$, the weight of $a \in \mathit{DA}^{+}(\omega) \cup \mathit{DA}^{-}(\omega)$ is $1$. Hence:
\[w(\omega) = \prod_{a \in \mathit{PA}^{+}(\omega)} p(a) \prod_{a \in \mathit{PA}^{-}(\omega)} (1-p(a)) \prod_{a \in \mathit{DA}^{+}(\omega)} 1 \prod_{a \in \mathit{DA}^{-}(\omega)} 1\]
\[= \prod_{a \in \mathit{PA}^{+}(\omega)} p(a) \prod_{a \in \mathit{PA}^{-}(\omega)} (1-p(a)) = P(\omega) .\]
\end{itemize}
This proves weight equivalence. $\Box$


\subsection*{Proof of Theorem 3}

\begin{theorem} 
Let $L_g$ be the relevant ground program for some ProbLog program with respect to $\mathbf{Q}$ and $\mathbf{E}=\mathbf{e}$. Let $\mathcal{M}$ be the corresponding ground MLN. The distribution $P(\mathbf{Q})$ according to $\mathcal{M}$ is the same as the distribution $P(\mathbf{Q} \mid \mathbf{E}=\mathbf{e})$ according to $L_g$.
\end{theorem}

\noindent \emph{Proof:} We prove that \textbf{(1)} the set of worlds with non-zero probability according to the MLN is the same as the set of worlds with non-zero probability according to the ProbLog program \emph{and} the evidence; \textbf{(2)} for every such world $\omega$, $P_{\mathcal{M}}(\mathbf{\omega})=P_L(\mathbf{\omega} \mid \mathbf{E}=\mathbf{e})$

\noindent \textbf{(Part 1)} A world has non-zero probability according to an MLN if it satisfies all hard clauses in the MLN. The hard clauses in the MLN are the same as the clauses in the weighted formula $\varphi$. Hence the set of worlds with non-zero probability according to the MLN equals $SAT(\varphi)$. Theorem 2 (model equivalence) implies that this set equals $MOD_{\mathbf{E}=\mathbf{e}}(L_g)$, which is exactly the set of worlds with non-zero probability according to the ProbLog program and the evidence.

\noindent \textbf{(Part 2)} The probability of a world $\omega \in SAT(\varphi)$ according to an MLN is defined as $P_{\mathcal{M}}(\mathbf{Q})=W(\omega)/Z$, with $W(\omega)$ the product of exponentiated weights of the soft clauses satisfied in $\omega$, and $Z$ the normalization constant. The probability of $\omega$ according to the ProbLog program conditioned on the evidence is $P_L(\omega | \mathbf{E}=\mathbf{e}) = P_L(\omega)/P_L(\mathbf{E}=\mathbf{e}$). We now show that both expressions are the same (i.e.\ $W(\omega)/Z=P_L(\omega)/P_L(\mathbf{E}=\mathbf{e})$). 
\begin{itemize}
\item The \underline{numerators} are the same ($W(\omega)=P_L(\omega)$): The only soft clauses in the MLN are unit clauses, whose weights are derived from the probabilistic facts. The unit clauses are such that, for any given world $\omega$, there is one unit clause per probabilistic atom that is satisfied. $W(\omega)$ is the product of the exponentiated weights of all these clauses. It follows from the way these weights are defined in terms of the weighted formula, and from weight equivalence between the weighted formula and the ProbLog program (Theorem 2), that this product is equal to the probability of the total choice of $\omega$ according to the ProbLog program and hence to the numerator $P_L(\omega)$. 
\item The \underline{denominators} are the same ($Z=P_L(\mathbf{E}=\mathbf{e})$): The normalization constant $Z$ of the MLN is defined as $\sum_{\omega \in SAT(\varphi)} W(\omega)$. The evidence probability $P_L(\mathbf{E}=\mathbf{e})$ equals $\sum_{\omega \in MOD_{\mathbf{E}=\mathbf{e}}(L)} P(\omega)$. These sums are equal since (a) the sets over which they range are equal due to Theorem 2 (model equivalencce), (b) the summed terms are equal (because of the same reasoning as for the numerator). 
\end{itemize}
This concludes the proof. $\Box$


\section{Markov Logic}
\label{app:MLN}

We briefly review Markov Logic \cite{MLN_PILPbook}. While Markov Logic generally works with FOL formulas, we consider only the ground case, as this is sufficient for our paper. 

A \emph{Markov Logic Network (MLN)} consists of two parts: a set of `soft' formulas $f_i$, which each have an associated weight $w_i \in \mathbb{R}$, and a set of `hard' formulas. An MLN determines a probability distribution on the set of possible worlds (determined by the Herbrand base). The probability of a world $\omega$ is 0 if it violates some hard formula and is $\frac{1}{Z}e^{\sum_i w_i \delta_i(\omega)}$ otherwise, where the sum is over all soft formulas and $\delta_i(\omega)$ is the indicator function being 1 if the soft formula $f_i$ is true in world $\omega$ and 0 otherwise.  Note that the exponent $\sum_i w_i \delta_i(\omega)$ is the sum of weights of satisfied soft formulas in world $\omega$; the higher this sum, the more likely $\omega$ is. The name `MLN' comes from the fact that this probability distribution can also be written as the distribution of a Markov network.



\section{The need for smoothing d-DNNFs}
\label{app:circuits}

 The algorithm we use to compute marginal probabilities requires a smooth d-DNNF. A smooth d-DNNF 
is a d-DNNF where for every disjunction node all children use exactly the same set of atoms. That is, if $C_1..C_n$ are the children of an $OR$ node $C$, then $Atoms(C_i) = Atoms(C_j)$,  for $i \neq j$, where $Atoms(C_i)$ is the set of atoms which $C_i$ uses. 
  
  \begin{figure}[t]
    \centering
    \subfigure[A non smooth d-DNNF]
    {\includegraphics[width=0.45\textwidth]{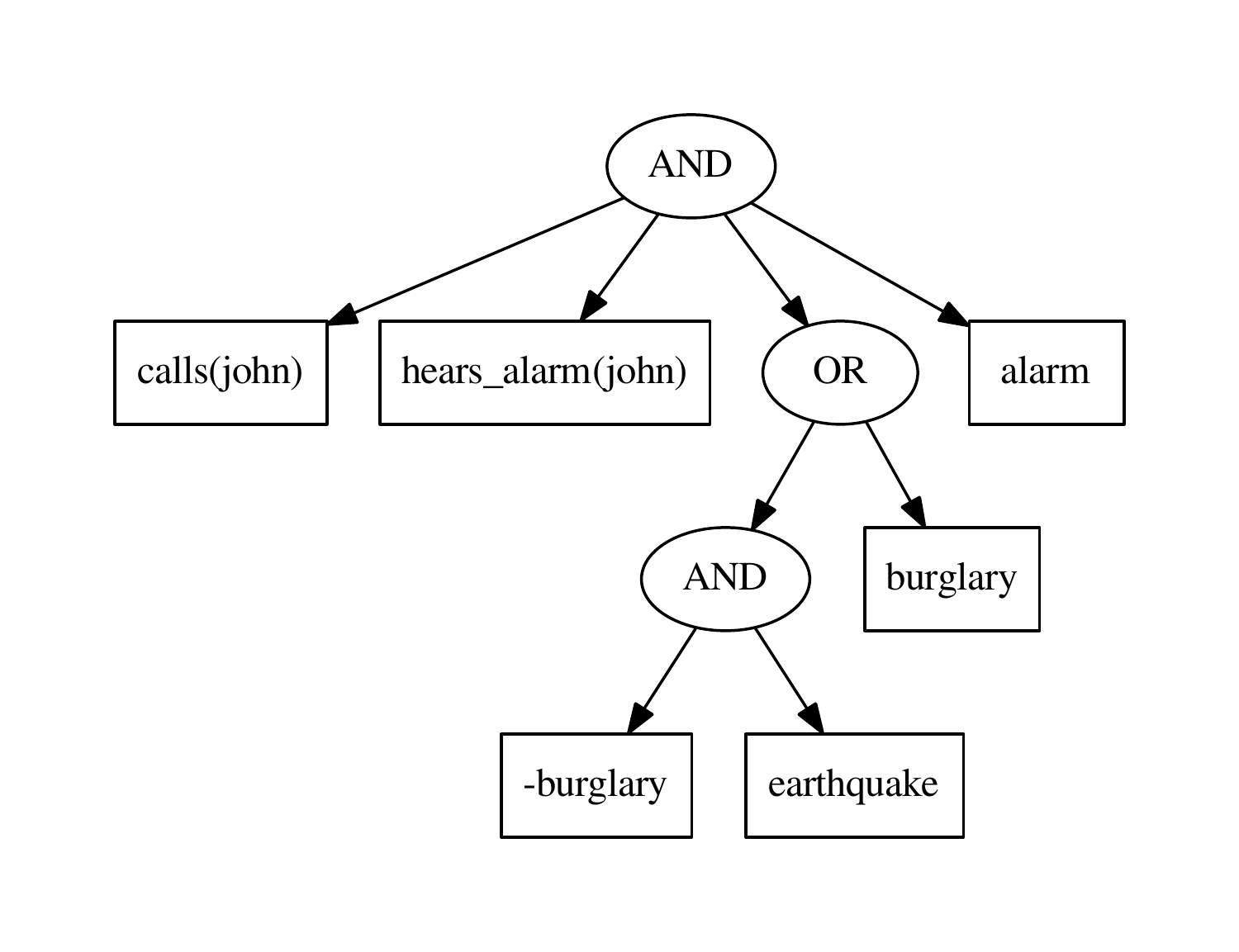}}
    \subfigure[A smooth d-DNNF]
    {\includegraphics[width=0.45\textwidth]{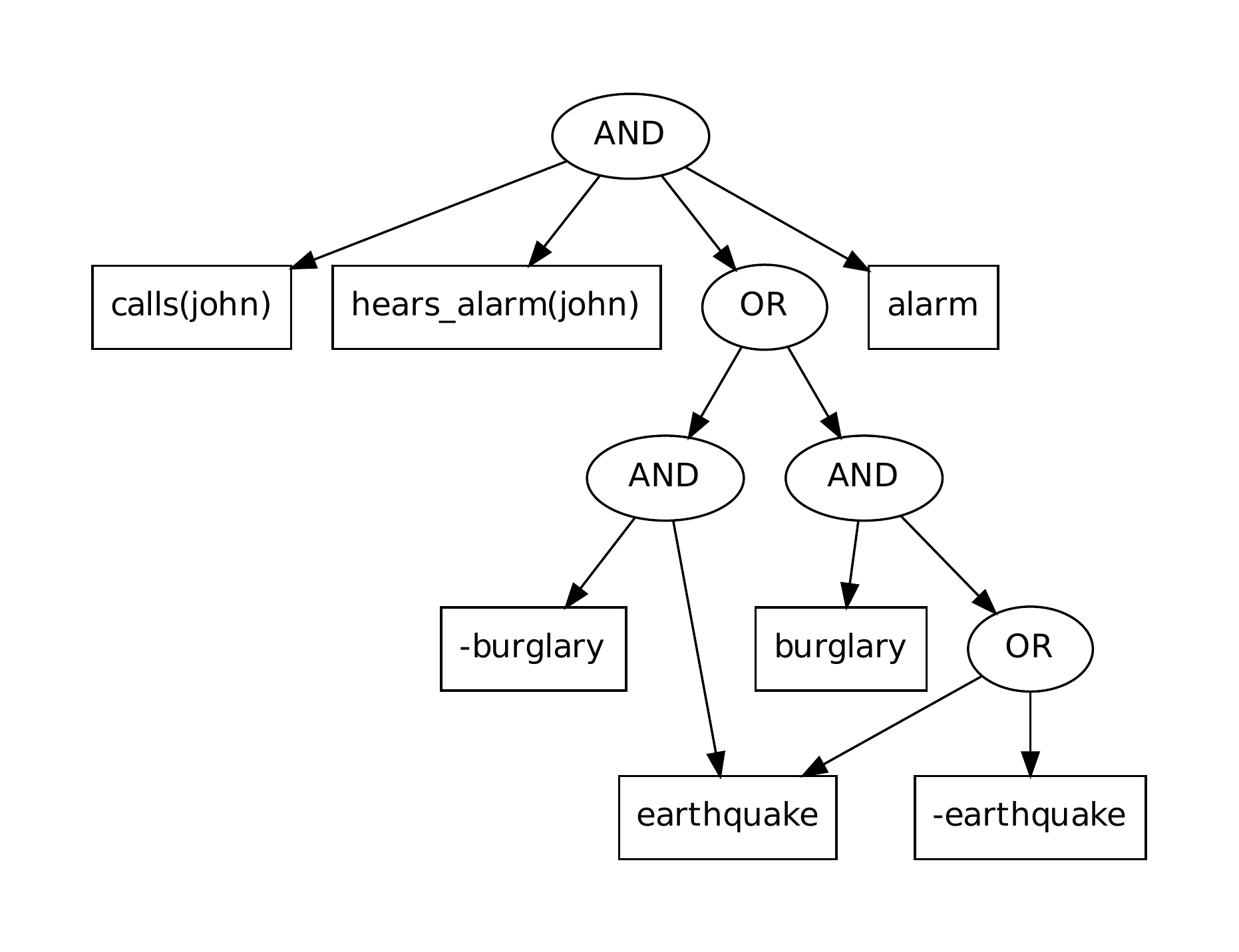}}
    \caption{A d-DNNF and the corresponding smooth d-DNNF for the formula $burglary \lor earthquake$.}
    \label{fig:app:ddnnfs_for_burglary_or_earthquake}
  \end{figure}
  
Figure~\ref{fig:app:ddnnfs_for_burglary_or_earthquake}a shows a non-smooth d-DNNF, while Figure~\ref{fig:app:ddnnfs_for_burglary_or_earthquake}b shows the corresponding smooth d-DNNF (which we have already shown before in Figure~\ref{fig:example5_ddnnf}a but we repeat here for convenience). Consider the non-smooth d-DNNF. The \verb|OR| node doesn't satisfy smoothness, since the sets of atoms of its children differ ($\{burglary,$ $earthquake\}$ and $\{burglary\}$; the negation is ignored here). Hence we need to transform this d-DNNF into a smooth d-DNNF, see Figure~\ref{fig:app:ddnnfs_for_burglary_or_earthquake}b. This is done by substituting the \verb|burglary| node by an \verb|AND| node, then adding the \verb|burglary| node as a child to the new \verb|AND| node and creating a new smoothing node for the missing atom \verb|-earthquake|. The smoothing node is an \verb|OR| node which links to \verb|earthquake| and \verb|-earthqake|. It is then linked to the \verb|AND| node. 
  
\begin{figure}[htb]
    \centering
    \includegraphics[width=1\textwidth]{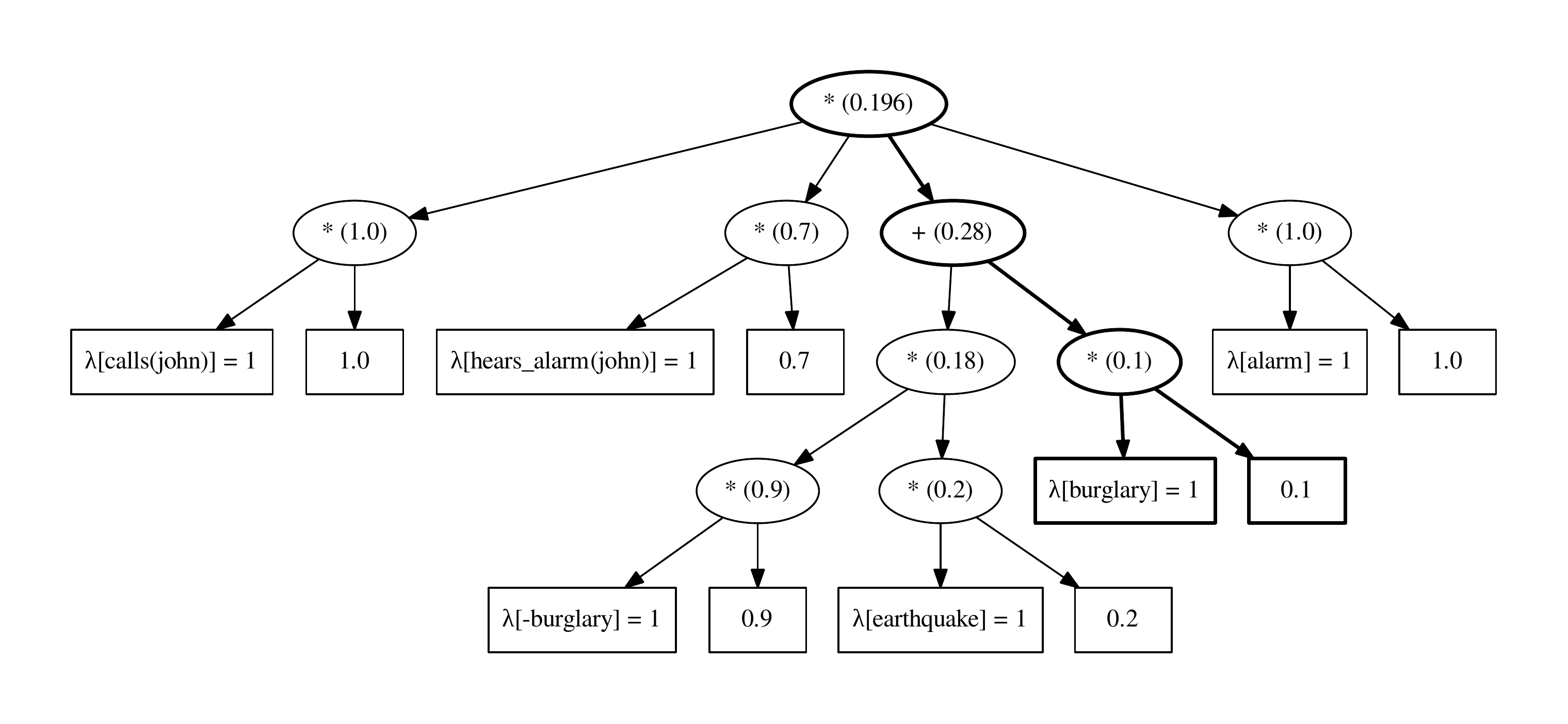}
    \caption{The arithmetic circuit corresponding to the non-smooth d-DNNF for the Alarm example.}
    \label{fig:app:ac_eval_smooth_vs_nosmooth}
  \end{figure}

Let us illustrate how smoothness affects the computation of probabilities using our Alarm running example (so not the restricted version of the example considered above). We have seen the arithmetic circuit (AC) corresponding to the smooth d-DNNF for this example before, recall Figure~\ref{fig:ACeval} on p.~\pageref{fig:ACeval}. This figure also illustrates how we can compute the probability of the conjunction $P(\mathit{earthquake} = \mathit{true} \land \mathit{calls}(\mathit{john}) =\mathit{true})$. This yields the value 0.14, which is indeed the correct value. In contrast, Figure~\ref{fig:app:ac_eval_smooth_vs_nosmooth} shows the same evaluation process on an AC for the non-smooth d-DNNF. This results in an incorrect value (0.196). This shows the need for smoothness of the d-DNNF. 

    
\section{Kullback-Leibler Divergence Between ProbLog Programs}
\label{apd:kldiv}
The Kullback-Leibler divergence $D(P||Q)$ is a non-symmetric measure for the difference of two probability distributions $P$ and $Q$ (cf.~\citeN{wasserman04allof}). It is used in probability theory as well as in information theory where it is also known as information gain.  The K-L divergence aggregates the difference of the two distributions on all elements of the outcome space. It is only defined if the support of $Q$ is larger than the one of $P$, that is, for all
$i$ where $P(i)>0$ also $Q(i)>0$.

We use the K-L divergence to evaluate the LFI-ProbLog learning algorithm (cf.~Algorithm~\ref{alg:lfi:mainloop}) and measure how close the learned program $T_2$ is to the ground truth program $T_1$.  We are doing parameter estimation, that is, the structure of the program is fixed and only the fact probabilities change. Hence we can restrict the definition of the K-L divergence to programs that are identical except for the fact probabilities.

\begin{definition}[K-L Divergence]
Let $T_1=F_1 \cup R$ and $T_2=F_2\cup R$ be ground ProbLog programs such that the probabilistic facts are identical except for the probabilities, that is, $F_1=\lbrace p_i :: f_i | 1\le i \le n\rbrace$ and $F_2=\lbrace q_i :: f_i | 1\le i \le n\rbrace$. Let $At$ denote the Herbrand base of $T_1$ and $T_2$ (note that they have the same Herbrand base). We denote interpretations as subsets of atoms, i.e., $L \subseteq At$ is the interpretation in which the atoms that are in $L$ are true and the other atoms  are false. Then the K-L Divergence between $T_1$ and $T_2$ is defined as
\begin{equation}
  D(T_1 || T_2) = \sum\limits_{L\subseteq At}
  P_{T_1}(L) \log\frac{P_{T_1}(L)}{P_{T_2}(L)}
\end{equation}
\end{definition}

There are exponentially many interpretations $L\subseteq At$, which makes evaluating the K-L divergence as defined above impossible in practice.  However, the probabilistic facts in a ProbLog program are independent, which can be exploited to compute the K-L divergence in linear time by looping once over $F$.

\begin{theorem}
  \label{theorem:appendix:kldiv}
  Let $T_1=F_1 \cup R$ and $T_2=F_2\cup R$ be ground ProbLog programs such that the probabilistic facts are identical except for the probabilities, that is, $F_1=\lbrace p_i :: f_i | 1\le i \le n\rbrace$ and $F_2=\lbrace q_i :: f_i | 1\le i \le n\rbrace$. Then the K-L Divergence between $T_1$ and $T_2$ can be calculated as
  \begin{equation}
    D(T_1 || T_2) = \sum\limits_{i=1}^n \left( p_i\log\frac{p_i}{q_i}+(1-p_i)\log\frac{1-p_i}{1-q_i} \right) \enspace .
  \end{equation}
\end{theorem}

It is possible to extend the K-L divergence and the theorem to
non-ground facts. To do so, one needs to multiply each summand
$p_i\log\frac{p_i}{q_i}+(1-p_i)\log\frac{1-p_i}{1-q_i}$ with the
number of ground instances of the probabilistic fact $f_i$.

\begin{proof}
  We prove Theorem~\ref{theorem:appendix:kldiv} by induction over the number of probabilistic facts.
  
  \textbf{Base case} $n=1$.
  \begin{eqnarray*}
    D(T_1 || T_2) & = & \sum\limits_{L\subseteq At} P_{T_1}(L)  \log\frac{P_{T_1}(L)}{P_{T_2}(L)} \\
    & = & P_{T_1}(\lbrace f_1\rbrace) \log\frac{P_{T_1}(\lbrace f_1\rbrace)}{P_{T_2}(\lbrace f_1\rbrace) }\\
    & & + P_{T_1}(\emptyset) \log\frac{P_{T_1}(\emptyset) }{P_{T_2}(\emptyset) }\\
    & = & p_1 \log\frac{p_1}{q_1}+ (1-p_1) \log\frac{1-p_1}{1-q_1}\\
    & = & \sum\limits_{i=1}^n \left(p_i\log\frac{p_i}{q_i}+(1-p_i)\log\frac{1-p_i}{1-q_i} \right)
  \end{eqnarray*}

  \textbf{Inductive case} $n\to n+1$. To simplify the notation, we define $T_1^{n+1} = T_1\cup \lbrace p_{n+1} :: f_{n+1}\rbrace$ and $T_2^{n+1} = T_2\cup \lbrace q_{n+1} :: f_{n+1}\rbrace$
  \begin{eqnarray*}
    D(T_1^{n+1} || T_2^{n+1}) & \\
    = &\sum\limits_{L\subseteq (At \cup \lbrace f_{n+1}\rbrace )} P_{T_1^{n+1}}(L) \log\frac{P_{T_1^{n+1}}(L)}{P_{T_2^{n+1}}(L)} \\
     =&\left[\sum\limits_{L\subseteq At }P_{T_1^{n+1}}(L\cup\lbrace f_{n+1}\rbrace) \log\frac{P_{T_1^{n+1}}(L\cup\lbrace f_{n+1}\rbrace)}{P_{T_2^{n+1}}(L\cup\lbrace f_{n+1}\rbrace)} \right]+\\
     &\left[\sum\limits_{L\subseteq At} P_{T_1^{n+1}}(L) \log\frac{P_{T_1^{n+1}}(L)}{P_{T_2^{n+1}}(L)} \right] \\
     & \textrm{Probabilistic facts are independent and thus we can} \\
     & \textrm{factorize the probabilities} \\
     =&\left[\sum\limits_{L\subseteq At }p_{n+1}\cdot P_{T_1}(L) \log\frac{p_{n+1}\cdot P_{T_1}(L)}{q_{n+1}\cdot P_{T_2}(L)}  \right]+\\
     &\left[\sum\limits_{L\subseteq At} (1-p_{n+1})\cdot P_{T_1}(L) \log\frac{(1-p_{n+1})\cdot P_{T_1}(L)}{(1-q_{n+1})\cdot P_{T_2}(L)} \right] \\
     &\textrm{using the rules for $\log$ and factoring out the constants} \\
    =& p_{n+1} \left[\sum\limits_{L\subseteq At } P_{T_1}(L) \left(\log\frac{p_{n+1}}{q_{n+1}}+\log\frac{ P_{T_1}(L)}{ P_{T_2}(L)}\right) \right]+\\
     &(1-p_{n+1}) \left[\sum\limits_{L\subseteq At } P_{T_1}(L) \left(\log\frac{1-p_{n+1}}{1-q_{n+1}}+\log\frac{ P_{T_1}(L)}{ P_{T_2}(L)}\right) \right] \\
     & \textrm{expanding the inner sums and factoring out constants} \\
    =& p_{n+1} \left(   \log\frac{ p_{n+1}}{ q_{n+1}} \right) \left[\sum\limits_{L\subseteq At }  P_{T_1}(L)\right]+\\
     &p_{n+1} \left[\sum\limits_{L\subseteq At }   P_{T_1}(L) \left(  \log\frac{   P_{T_1}(L)}{   P_{T_2}(L)}\right) \right]+\\
     &(1-p_{n+1}) \left( \log\frac{1-p_{n+1}}{1-q_{n+1}} \right) \left[\sum\limits_{L\subseteq At}   P_{T_1}(L)\right]+ \\
     &(1-p_{n+1})\left[\sum\limits_{L\subseteq At}   P_{T_1}(L) \left( \log\frac{  P_{T_1}(L)}{  P_{T_2}(L)} \right) \right] \\
     & \textrm{since $\sum_{L\subseteq At }   P_{T_1}(L)$ is 1, rearranging yields}\\ 
     =& p_{n+1} \left(   \log\frac{ p_{n+1}}{ q_{n+1}} \right) +(1-p_{n+1}) \left( \log\frac{1-p_{n+1}}{1-q_{n+1}} \right)+\\
     &\sum\limits_{L\subseteq At }   P_{T_1}(L) \log\frac{   P_{T_1}(L)}{   P_{T_2}(L)} \\
     &\textrm{using the inductive assumption}\\
     =& p_{n+1} \left(   \log\frac{ p_{n+1}}{ q_{n+1}} \right) +(1-p_{n+1}) \left( \log\frac{1-p_{n+1}}{1-q_{n+1}} \right)+\\
     & \sum\limits_{i=1}^n  \left(p_i\log\frac{p_i}{q_i}+(1-p_i)\log\frac{1-p_i}{1-q_i}\right) \\
     & \textrm{rearranging the terms}\\
     =&\sum\limits_{i=1}^{n+1}  \left(p_i\log\frac{p_i}{q_i}+(1-p_i)\log\frac{1-p_i}{1-q_i} \right)
  \end{eqnarray*}
\end{proof}

\section*{Acknowledgements} 

We thank the reviewers for their useful suggestions.  We thank Maurice Bruynooghe, Jesse Davis, Kristian Kersting, Angelika Kimmig and Theofrastos Mantadelis for useful discussions.

Daan Fierens, Guy Van den Broeck and Bernd Gutmann are supported by the Research Foundation-Flanders (FWO-Vlaanderen). Joris Renkens is supported by PF-10/010 NATAR. Research supported by the European Commission under contract number FP7-248258-First-MM.

\bibliography{plp2cnf.bib}

\end{document}